\pdfoutput=1
\documentclass[12pt]{article}
\usepackage[T1]{fontenc}
\usepackage[utf8]{inputenc}
\usepackage{tocloft}
\usepackage[height=8.85in,width=6.45in]{geometry}
\usepackage[skip=1em plus 0.2em minus 0.1em]{parskip}
\usepackage{amsmath,amssymb,mathtools}
\usepackage{slashed,braket}
\usepackage{graphicx}
\usepackage{multirow,adjustbox}
\usepackage{subcaption}
\usepackage{enumitem}
\usepackage[svgnames]{xcolor}
\usepackage[colorlinks,citecolor=DarkGreen,linkcolor=FireBrick,
  urlcolor=FireBrick,linktocpage,unicode]{hyperref}
\usepackage[hang,flushmargin,bottom]{footmisc}
\PassOptionsToPackage{disable}{todonotes}
\allowdisplaybreaks
\usepackage{graphicx}
\usepackage{tikz}
\usepackage{tikz-cd}
\usepackage{times}
 
\usepackage{bm}
\usepackage{physics}
\usepackage{xcolor}
\usepackage{natbib}
\usepackage{mdframed}
\usepackage{nicefrac}
\usepackage{booktabs}
\usepackage{lipsum}
\usepackage{titlesec}
\usepackage{wrapfig,lipsum,booktabs}
\usepackage{authblk}
\usepackage{blindtext}

\usepackage{algorithm}
\usepackage{algpseudocode}
\newcommand{\commentsymbol}{//}\algrenewcommand\algorithmiccomment[1]{\hfill {\footnotesize \commentsymbol{} #1}}

\usepackage{titletoc}
\usepackage{todonotes}
\usepackage{authblk}
\usepackage{setspace}
\usepackage{dsfont} \newcommand{\one}{\mathds{1}}

\usepackage[nameinlink,capitalize,noabbrev]{cleveref}

\usepackage{CJK}

\usepackage{array}
\usepackage{bbm}
\usepackage{makecell}

\usepackage{dashbox}
\usepackage{xcolor}
\usepackage{colortbl}
\definecolor{lightyellow}{rgb}{1.0, 0.95, 0.7}
\definecolor{Blue}{rgb}{0, 0, 0.8}
\definecolor{blue}{rgb}{0,0,1}
\definecolor{darkgreen}{rgb}{0,0.40,0}
\definecolor{firebrick}{rgb}{0.698,0.133,0.133}

\definecolor{colorA}{rgb}{1,0,0}
\definecolor{colorB}{rgb}{0,0.3,1}
\definecolor{colorC}{rgb}{0.9,0.8,0.2}
\definecolor{colorD}{rgb}{0,0.65,0}
\definecolor{lesslightgray}{rgb}{0.5,0.5,0.5}
\definecolor{light-gray}{gray}{0.95}

\let\tilde\widetilde

\def\P{\mathbb{P}}
\def\R{\mathbb{R}}

\DeclareMathOperator*{\E}{{\mathbb{E}}} 

\let\cite\citep 

\usepackage{amsthm}
\usepackage[many]{tcolorbox}

\usepackage{etoolbox}

\newtheoremstyle{theoremstyle}
  {.5\baselineskip} {.5\baselineskip} {}                  {}                  {\bfseries}        {.}                 {1em}               {}

\theoremstyle{theoremstyle}
\newtheorem{theorem}{Theorem}[section]
\newtheorem{lemma}{Lemma}[section]

\newtheorem{proposition}{Proposition}[section]
\newtheorem{definition}{Definition}[section]

\newtheorem{remark}{Remark}[section]

\tcolorboxenvironment{theorem}{
  breakable,
  colback=black!10,
  colframe=white,width=\dimexpr\linewidth+10pt\relax,enlarge left by=-5pt,enlarge right by=-5pt,boxsep=5pt,boxrule=0pt,
  left=0pt,right=0pt,top=0pt,bottom=0pt,
  sharp corners,
  before skip=0.5\baselineskip, after skip=0.5\baselineskip,  fonttitle=\bfseries, coltitle=black }
\tcolorboxenvironment{remark}{
  blanker,
  breakable,
  before skip=.8\baselineskip,  after  skip=.8\baselineskip   }

\tcolorboxenvironment{proposition}{
  breakable,
  colback=black!10,
  colframe=white,width=\dimexpr\linewidth+10pt\relax,enlarge left by=-5pt,enlarge right by=-5pt,boxsep=5pt,boxrule=0pt,
  left=0pt,right=0pt,top=0pt,bottom=0pt,
  sharp corners,
  before skip=0.5\baselineskip, after skip=0.5\baselineskip,  fonttitle=\bfseries, coltitle=black }

\tcolorboxenvironment{lemma}{
  breakable,
  colback=black!10,
  colframe=white,width=\dimexpr\linewidth+10pt\relax,enlarge left by=-5pt,enlarge right by=-5pt,boxsep=5pt,boxrule=0pt,
  left=0pt,right=0pt,top=0pt,bottom=0pt,
  sharp corners,
  before skip=0.5\baselineskip, after skip=0.5\baselineskip,  fonttitle=\bfseries, coltitle=black }

\tcolorboxenvironment{corollary}{
  breakable,
  colback=black!10,
  colframe=white,width=\dimexpr\linewidth+10pt\relax,enlarge left by=-5pt,enlarge right by=-5pt,boxsep=5pt,boxrule=0pt,
  left=0pt,right=0pt,top=0pt,bottom=0pt,
  sharp corners,
  before skip=0.5\baselineskip, after skip=0.5\baselineskip,  fonttitle=\bfseries, coltitle=black }

\tcolorboxenvironment{definition}{
  breakable,
  colback=black!10,
  colframe=white,width=\dimexpr\linewidth+10pt\relax,enlarge left by=-5pt,enlarge right by=-5pt,boxsep=5pt,boxrule=0pt,
  left=0pt,right=0pt,top=0pt,bottom=0pt,
  sharp corners,
  before skip=0.5\baselineskip, after skip=0.5\baselineskip,  fonttitle=\bfseries, coltitle=black }
\tcolorboxenvironment{assumption}{
  breakable,
  colback=black!10,
  colframe=white,width=\dimexpr\linewidth+10pt\relax,enlarge left by=-5pt,enlarge right by=-5pt,boxsep=5pt,boxrule=0pt,
  left=0pt,right=0pt,top=0pt,bottom=0pt,
  sharp corners,
  before skip=0.5\baselineskip, after skip=0.5\baselineskip,  fonttitle=\bfseries, coltitle=black }

\crefname{theorem}{Theorem}{Theorems}
\crefname{proposition}{Proposition}{Propositions}
\crefname{lemma}{Lemma}{Lemmas}
\crefname{corollary}{Corollary}{Corollaries}
\crefname{definition}{Definition}{Definitions}
\crefname{assumption}{Assumption}{Assumptions}
\crefname{remark}{Remark}{Remarks}
\crefname{problem}{Problem}{Problems}
\crefname{property}{Property}{property}

\tcolorboxenvironment{hypothesis}{
  breakable,
  colback=black!10,
  colframe=white,width=\dimexpr\linewidth+10pt\relax,enlarge left by=-5pt,enlarge right by=-5pt,boxsep=5pt,boxrule=0pt,
  left=0pt,right=0pt,top=0pt,bottom=0pt,
  sharp corners,
  before skip=0.5\baselineskip, after skip=0.5\baselineskip,  fonttitle=\bfseries, coltitle=black }
\crefname{hypothesis}{Hypothesis}{Hypothesises}

\tcolorboxenvironment{fact}{
  breakable,
  colback=black!10,
  colframe=white,width=\dimexpr\linewidth+10pt\relax,enlarge left by=-5pt,enlarge right by=-5pt,boxsep=5pt,boxrule=0pt,
  left=0pt,right=0pt,top=0pt,bottom=0pt,
  sharp corners,
  before skip=0.5\baselineskip, after skip=0.5\baselineskip,  fonttitle=\bfseries, coltitle=black }
\crefname{fact}{Fact}{Facts}

\tcolorboxenvironment{example}{
  breakable,
  colback=black!10,
  colframe=white,width=\dimexpr\linewidth+10pt\relax,enlarge left by=-5pt,enlarge right by=-5pt,boxsep=5pt,boxrule=0pt,
  left=0pt,right=0pt,top=0pt,bottom=0pt,
  sharp corners,
  before skip=0.5\baselineskip, after skip=0.5\baselineskip,  fonttitle=\bfseries, coltitle=black }
\crefname{example}{Example}{Examples}

\tcolorboxenvironment{question}{
  breakable,
  colback=black!10,
  colframe=white,width=\dimexpr\linewidth+10pt\relax,enlarge left by=-5pt,enlarge right by=-5pt,boxsep=5pt,boxrule=0pt,
  left=0pt,right=0pt,top=0pt,bottom=0pt,
  sharp corners,
  before skip=0.5\baselineskip, after skip=0.5\baselineskip,  fonttitle=\bfseries, coltitle=black }

\crefname{question}{Question}{Questions}
\numberwithin{equation}{section}
\numberwithin{theorem}{section}
\numberwithin{proposition}{section}
\numberwithin{definition}{section}
\numberwithin{lemma}{section}
\numberwithin{assumption}{section}
\numberwithin{remark}{section}

\usepackage{lipsum}

\newcommand*{\annot}[1]{\tag*{\footnotesize{\textcolor{black!50}{\big(#1\big)}}}}

\makeatletter
\let\save@mathaccent\mathaccent
\newcommand*\if@single[3]{\setbox0\hbox{${\mathaccent"0362{#1}}^H$}\setbox2\hbox{${\mathaccent"0362{\kern0pt#1}}^H$}\ifdim\ht0=\ht2 #3\else #2\fi
}
\newcommand*\rel@kern[1]{\kern#1\dimexpr\macc@kerna}
\newcommand*\widebar[1]{\@ifnextchar^{{\wide@bar{#1}{0}}}{\wide@bar{#1}{1}}}
\newcommand*\wide@bar[2]{\if@single{#1}{\wide@bar@{#1}{#2}{1}}{\wide@bar@{#1}{#2}{2}}}
\newcommand*\wide@bar@[3]{\begingroup
    \def\mathaccent##1##2{\let\mathaccent\save@mathaccent
        \if#32 \let\macc@nucleus\first@char \fi
        \setbox\z@\hbox{$\macc@style{\macc@nucleus}_{}$}\setbox\tw@\hbox{$\macc@style{\macc@nucleus}{}_{}$}\dimen@\wd\tw@
        \advance\dimen@-\wd\z@
        \divide\dimen@ 3
        \@tempdima\wd\tw@
        \advance\@tempdima-\scriptspace
        \divide\@tempdima 10
        \advance\dimen@-\@tempdima
        \ifdim\dimen@>\z@ \dimen@0pt\fi
        \rel@kern{0.6}\kern-\dimen@
        \if#31
        \overline{\rel@kern{-0.6}\kern\dimen@\macc@nucleus\rel@kern{0.4}\kern\dimen@}\advance\dimen@0.4\dimexpr\macc@kerna
        \let\final@kern#2\ifdim\dimen@<\z@ \let\final@kern1\fi
        \if\final@kern1 \kern-\dimen@\fi
        \else
        \overline{\rel@kern{-0.6}\kern\dimen@#1}\fi
    }\macc@depth\@ne
    \let\math@bgroup\@empty \let\math@egroup\macc@set@skewchar
    \mathsurround\z@ \frozen@everymath{\mathgroup\macc@group\relax}\macc@set@skewchar\relax
    \let\mathaccentV\macc@nested@a
    \if#31
    \macc@nested@a\relax111{#1}\else
    \def\gobble@till@marker##1\endmarker{}\futurelet\first@char\gobble@till@marker#1\endmarker
    \ifcat\noexpand\first@char A\else
    \def\first@char{}\fi
    \macc@nested@a\relax111{\first@char}\fi
    \endgroup
    }
\makeatother
\let\bar\widebar

\usepackage{paper_metadata}
\PaperTitle{Flow Matching Reinforcement Learning via~SDE~Inference}

\PaperAffiliation{nu_cs}{Department of Computer Science,
  Northwestern University, Evanston, IL 60208, USA}
\PaperAffiliation{nu_foundation_models}{Center for Foundation Models and Generative AI,
  Northwestern University, Evanston, IL 60208, USA}
\PaperAffiliation{ntu}{Department of  Electrical Engineering, National Taiwan University, Taipei, Taiwan}
\PaperAffiliation{nu_iems}{Department of Industrial Engineering and Management Sciences,\\
  Northwestern University, Evanston, IL 60208, USA}
\PaperAffiliation{nu_statistics}{Department of Statistics and Data Science,
  Northwestern University, Evanston, IL 60208, USA}

\PaperNote{equal}{These authors contributed equally. Project page: \url{https://magics-lab.github.io/flow_sde/}.}

\PaperAuthor[affiliations={nu_cs,nu_foundation_models},notes={equal}]{Maojiang Su}
\PaperAuthor[affiliations={ntu},notes={equal}]{Tsung-En Lin}
\PaperAuthor[affiliations={nu_cs,nu_foundation_models},notes={equal}]{Jerry Yao-Chieh Hu}
\PaperAuthor[affiliations={nu_cs}]{Guo Ye}
\PaperAuthor[affiliations={nu_cs}]{Haoran Lu}
\PaperAuthor[affiliations={nu_cs,nu_foundation_models}]{Shang Wu}
\PaperAuthor[affiliations={nu_cs,nu_iems}]{Zhaoran Wang}
\PaperAuthor[affiliations={nu_cs,nu_foundation_models,nu_statistics}]{Han Liu}

\usepackage[nopatch=footnote]{microtype}
\AtBeginEnvironment{algorithmic}{\footnotesize\setstretch{1}\setlength{\abovedisplayskip}{4pt}\setlength{\belowdisplayskip}{4pt}\setlength{\abovedisplayshortskip}{4pt}\setlength{\belowdisplayshortskip}{4pt}}

\hypersetup{
  pdftitle={Flow Matching Reinforcement Learning via SDE Inference},
  pdfauthor={Maojiang Su; Tsung-En Lin; Jerry Yao-Chieh Hu; Guo Ye;
    Haoran Lu; Shang Wu; Zhaoran Wang; Han Liu}
}
\setlist[itemize,enumerate]{
  parsep=\parskip,
  itemsep=\dimexpr .3em - \parskip\relax plus 2pt,
  topsep=\dimexpr 6pt - \parskip\relax plus 1pt minus 1pt,
  partopsep=0pt,
  listparindent=\parindent
}

\usepackage{etoolbox}
\newlength{\legacytheoremtopsep}
\newcommand{\localtheoremlistskips}{\setlength{\topsep}{\legacytheoremtopsep}\setlength{\partopsep}{\topsep}}
\makeatletter
\newcommand{\template@registermathspacing}[1]{\AddToHook{env/#1/before}[template-local-spacing]{\edef\template@outerlistdepth{\number\@listdepth}}\DeclareHookRule{env/#1/before}{template-local-spacing}{before}{tcolorbox}\AtBeginEnvironment{#1}{\ifnum\template@outerlistdepth=0\relax
      \localtheoremlistskips
    \fi
  }}
\forcsvlist{\template@registermathspacing}{theorem,lemma,corollary,proposition,definition,assumption,remark,problem,property,hypothesis,fact,example,question,claim}
\makeatother
\makeatletter
\newcommand{\template@titlelistskips}{\ifdefstring{\template@parentenv}{titlepage}{\localtheoremlistskips}{}}
\BeforeBeginEnvironment{center}{\let\template@parentenv\@currenvir}
\AtBeginEnvironment{center}{\template@titlelistskips}
\BeforeBeginEnvironment{flushright}{\let\template@parentenv\@currenvir}
\AtBeginEnvironment{flushright}{\template@titlelistskips}
\makeatother
\makeatletter
\expandafter\patchcmd\csname\string\proof\endcsname
  {\topsep6\p@\@plus6\p@\relax}
  {\topsep=3pt plus 1pt\relax\partopsep=0pt\relax}
  {}{\PackageError{proofspacing}{Could not patch amsthm proof}{}}
\makeatother

\begin{document}
\hypersetup{pageanchor=false}
\begin{titlepage}
\begin{flushright}
Last Update: November 18, 2026
\end{flushright}

\vskip 2.5em
\begin{center}
{\LARGE\bfseries
  \begin{spacing}{1.15}\PrintPaperTitle\end{spacing}}
\vskip 1em
\PrintPaperAuthors
\vskip 1em
{\setlength{\tabcolsep}{4pt}\PrintPaperAffiliations\par}
\PrintPaperNotes
\end{center}

\noindent
We develop a unified framework for flow-based reinforcement learning (RL) grounded in diffusion–flow duality and establish its theoretical foundations. 
The framework enables RL for deterministic flow matching inference by introducing diffusion stochastic inference dynamics that supports exploration while admitting deterministic deployment. 
We show that our framework subsumes several existing flow-based RL methods as special cases and inspire effective new methods. Then we provide theoretical foundations for the frameworks by establishing the correctness and performance transfer guarantee. Specifically, we prove that the policies optimized under stochastic dynamics close to deterministic dynamics at deployment when the pretrained model is well-trained. Moreover, we show that the policy improvement achieved under training-time SDE inference transfers to deployment-time ODE inference. 
Experiments on synthetic and continuous-control tasks validate the theory and demonstrate the proposed algorithms' effectiveness.

\end{titlepage}
\pagenumbering{arabic}
\hypersetup{pageanchor=true}

\begingroup
\setlength{\parskip}{0pt}
\setcounter{tocdepth}{2}
\tableofcontents
\endgroup

\section{Introduction}
\label{sec:intro}
Flow Matching (FM) \cite{lipman2022flow, albergo2023stochastic, liu2022flow} has become a powerful class of generative models and achieves state-of-the-art performance across image, speech, video generation, and robotics \cite{esser2024scaling, le2023voicebox, polyak2025moviegencastmedia, black2024pi}.
Motivated by this success, recent work explores Reinforcement Learning (RL) fine-tuning of flow-based models and policies to improve alignment, generation quality, and task-specific performance \cite{park2025flow, mcallister2025flow, fan2025online, wagenmaker2025steering}.

Growing interest in Stochastic Differential Equation (SDE) inference during FM training is motivated by its potential for effective exploration and policy gradient optimization.
Several recent empirical works illustrate this trend.
Flow-GRPO \cite{liu2025flow} introduces SDE-based inference for RL of flow matching models and achieves strong gains on text-to-image generation, including compositional generation and human preference alignment.
DanceGRPO \cite{xue2025dancegrpo} scales this paradigm to large-scale visual generation with improved stability across diverse prompts.
MixGRPO \cite{li2025mixgrpo} and TempFlow-GRPO \cite{he2025tempflow} further investigate partial or time-localized SDE inference by combining ODE and SDE sampling, leading to accelerated training.
In parallel, ReinFlow \cite{zhang2025reinflow} and Flow-SDE \cite{chen2025pi} apply SDE inference to flow policy optimization in robotics, demonstrating substantial gains under sparse rewards.

However, despite their strong empirical performance, these methods lack a unified framework that characterizes their common structure and theoretical guarantees for transforming deterministic ODE inference into stochastic SDE inference during training.
We address these gaps by introducing a unified framework for flow matching RL based on SDE inference.
The proposed framework formulates SDE inference as an inner Markov Decision Process (MDP) and applies policy gradient methods, such as PPO \cite{schulman2017proximal} and GRPO \cite{shao2024deepseekmath}, to optimize flow policies.
It subsumes existing flow-based RL methods
\cite{liu2025flow, xue2025dancegrpo, zhang2025reinflow,chen2025pi} as special cases and inspires new algorithmic instantiations.
Moreover, we present a theoretical analysis of correctness and RL performance transfer within this framework.
These theoretical results also apply to the aforementioned methods as special cases of our framework.

\textbf{Contributions.} Our contributions are threefold:
\begin{itemize}
    \item \textbf{Flow Matching RL Via Stochastic Inference. }
    We present a unified framework for flow matching RL based on stochastic SDE inference.
    For flow matching RL fine-tuning (\cref{sec:fm_rl_finetune}), with no external environment MDP, as in image generation, the framework applies policy gradient methods to optimize the flow matching model in the \textit{inner} MDP \eqref{eq:inner_mdp} induced by stochastic inference.
    For flow policy optimization (\cref{sec:fm_policy_op}), with an environment MDP, as in robotic control, the framework composes the inner inference and environment MDPs into a \textit{composite} MDP \eqref{eq:composite_mdp} and optimizes the flow-based policy in this composite MDP using policy gradient methods.
    Our framework subsumes existing flow-based RL algorithms through their surrogate objectives, noise levels and advantage estimators
    (\cref{sec:unified_framework}).
    Moreover, we introduce a new flow matching policy optimization algorithm 
    \textsc{FMPO-CPPO} (\cref{algo:FMPO_chainppo}), an instantiation of the proposed framework.
    
    \item \textbf{Theoretical Analysis.}
    We provide a rigorous theoretical analysis for flow matching RL via stochastic SDE inference.
    First, we establish a correctness guarantee (\cref{thm:correctness}), showing that when the pretrained flow model has small flow-matching error, stochastic SDE inference and deterministic ODE inference induce close terminal distributions.
    Second, we prove a performance transfer theorem (\cref{thm:improve_trans}), showing that policy improvement under training-time SDE inference transfers to deployment-time ODE inference up to an error term controlled by the flow-matching error.
    Together, these results give conditions for using stochastic SDE inference during training while preserving deterministic ODE inference at deployment.

    \item \textbf{Empirical Validation.}
    We experimentally validate the theoretical results and evaluate the proposed algorithms.
    Specifically, we validate the correctness theorem in \cref{sec:validate_corr} and the performance transfer theorem in \cref{sec:validate_trans}.
    The results confirm that smaller flow-matching error leads to smaller SDE-ODE mismatch and more reliable transfer from stochastic training to deterministic deployment.
    In addition, we evaluate \textsc{FMPO-CPPO} across representative continuous-control tasks in \cref{sec:algorithm_eva}, where it outperforms existing flow-based policy optimization baselines.
\end{itemize}

\vspace{-2mm}
\paragraph{Related Works.}
We defer extended related-work discussion to \cref{sec:related_works} due to page limits.

\vspace{-2mm}
\paragraph{Organization.}
\cref{sec:background} reviews flow matching and SDE inference.
\cref{sec:framework} presents a unified framework for flow-based RL under stochastic inference.
\cref{sec:theoretical_analysis} establishes its theoretical properties.
\cref{sec:exp} presents validation experiments and empirical results.
\cref{sec:conclusion} concludes with practical insights.
The appendix provides additional related work (\cref{sec:related_works}) and detailed proofs (\cref{sec:proof_correctness,sec:proof_improve_trans}).

\begin{table*}[t]
\centering
\caption{
Flow-based reinforcement learning methods under a unified stochastic inference perspective.
}
\label{tab:unified_framework}
\resizebox{1\columnwidth}{!}{\begin{tabular}{l|cccc}
\toprule
Method
& SDE Inference Horizon
& Surrogate
& Advantage Estimator
& Noise Level $\lambda_t$
\\
\midrule
\multicolumn{5}{l}{\textit{Flow Matching Reinforcement Learning (FMRL)}} \\
\midrule
Flow-GRPO \cite{liu2025flow}
& Full trajectory
& GRPO
& Group-relative 
& $a(1-t)/t$
\\

DanceGRPO \cite{xue2025dancegrpo}
& Full trajectory
& GRPO
& Group-relative 
& Constant
\\

MixGRPO \cite{li2025mixgrpo}
& Local window
& GRPO
& Group-relative 
& $a(1-t)/t$
\\

TempFlow-GRPO \cite{he2025tempflow}
& Time-localized
& GRPO
& Group-relative 
& $a(1-t)/t$
\\

\midrule
\multicolumn{5}{l}{\textit{Flow Matching Policy Optimization (FMPO)}} \\
\midrule
ReinFlow \cite{zhang2025reinflow}
& Full trajectory
& PPO
& GAE 
& Learnable
\\

Flow-SDE
\cite{chen2025pi}
& Full trajectory
& PPO
& GAE 
& $a(1-t)/t$
\\

FMPO-CPPO (ours)
& Full trajectory
& ChainPPO
& GAE 
& $a(1-t)$
\\

\bottomrule
\end{tabular}
}
\vspace{-1em}
\end{table*}

\section{Preliminaries}
\label{sec:background}
\paragraph{Flow Matching.}
A flow model is a continuous transformation that maps samples
$X_0 \sim p_0$ from a source distribution to samples
$X_1 \sim p_1$ from a target distribution.
A time-dependent flow $\psi: [0,1] \times \R^d \rightarrow \R^d$ implements this transformation, where $\psi_t(x)$ denotes the state of $x$ at time $t$.
The generative process $(X_t)_{t\in[0,1]}$ follows $X_0 \sim p_0$,
\begin{align*}
    X_t = \psi_t(X_0),\quad t \in [0,1].
\end{align*}
A velocity field $u:[0,1] \times \R^d \to \R^d$ defines a flow $\psi$ via an Ordinary Differential Equation (ODE):
\begin{align}
\label{eqn:flow_ODE}
    \dv{\psi_t}{t} = u(\psi_t(x),t), \quad
    \psi_0(x) = x.
\end{align}
This ODE induces a  probability path
$(p_t)_{t\in[0,1]}$ for the marginal distribution
$X_t=\psi_t(X_0)$.

Flow Matching \cite{lipman2022flow} learns a parameterized velocity field $u_\theta$
to match the pre-designed velocity $u$ inducing the desired probability path $p_t$ from $p_0$ to $p_1$.
Define the flow matching loss 
\begin{align}
\label{eqn:fm_loss_1}
    \mathcal{L}_{\text{FM}}(\theta) = \E_{t,X_t \sim p_t}[\|u_{\theta}(X_t,t)-u(X_t,t)\|_2^2], 
\end{align}
where $t \sim U[0,1]$.
However, the target velocity $u(x,t)$ is intractable.
To resolve this issue, \citet{lipman2022flow} introduce an equivalent conditional flow matching loss
\begin{align}
    \mathcal{L}_{\text{CFM}}(\theta) 
    =
    \E_{t,X_1, X_t}
    \Big[
        \big\|
            u_{\theta}(X_t, t)
            -
            u(X_t|X_1, t)
        \big\|_2^2
    \Big],
    \label{eq:con_fm_loss_2}
\end{align}
where $X_1 \sim p_1$ and $X_t \sim p_{t|1}(\cdot|X_1)$.
For a fixed probability path, coupling, and time distribution,
$\mathcal{L}_{\mathrm{CFM}}(\theta)=\mathcal{L}_{\mathrm{FM}}(\theta)+C$,
where $C$ is independent of $\theta$; hence the parameter gradients and minimizers coincide.
Following \cite{lipman2022flow,lipman2024flow},
we use the affine path
$ X_t = \alpha_t X_1 + \sigma_t X_0$.
Its ground-truth conditional velocity is
$u(x | x_1,t)=\dot{\alpha}_t x_1+(\dot{\sigma}_t/\sigma_t)(x-\alpha_t x_1)$ for $\sigma_t>0$. For the linear path, this becomes $u(x|x_1,t)=(x_1-x)/(1-t)$ for $t<1$.
The schedules $\alpha_t,\sigma_t:[0,1]\to[0,1]$ are smooth and monotone functions satisfying
$\alpha_0=\sigma_1=0$ and $\alpha_1=\sigma_0=1$.
After training $u_\theta$,
we sample by solving \eqref{eqn:flow_ODE} with $u$ replaced by $u_\theta$; the terminal distribution approximates $p_1$.

\textbf{Flow Matching SDE Inference.}
For any noise level $\lambda_t \geq 0$, we consider two stochastic processes 
initialized from the same distribution $p_0$
\begin{align}
\dd X_t
= & ~ u(X_t,t) \dd t
\label{eq:ode_process}\\
\dd \Tilde{X}_t
= & ~ [u(\Tilde{X}_t,t)+\frac{\lambda_t}{2} \nabla \log p_t(\Tilde{X}_t)] \dd t + \sqrt{\lambda_t} \dd W_t, 
\label{eq:sde_process}
\end{align}
where $W_t$ is the Wiener process.
\eqref{eq:ode_process} corresponds to the standard ODE inference used in flow matching.
Then both processes induce the same marginal distributions $p_t$ for all $t \in [0,1]$.
For the affine path with $X_0\sim\mathcal{N}(0,I)$ independent of $X_1$, \citet{lipman2024flow} derive the following score-based representation of the velocity at times with $\alpha_t\ne0$:
\begin{align}
\label{eq:velocity_score}
    u(x,t) = \frac{\dot{\alpha}_t}{\alpha_t} x 
    - \frac{\dot{\sigma}_t \sigma_t \alpha_t - \dot{\alpha}_t \sigma_t^2}{\alpha_t} 
    \nabla \log p_t(x).
\end{align}
Substituting \eqref{eq:velocity_score} into \eqref{eq:sde_process} yields the following SDE with $\Tilde{X}_0 \sim p_0$ at times where $\dot{\sigma}_t\sigma_t\alpha_t-\dot{\alpha}_t\sigma_t^2\ne0$; endpoint coefficients are defined by limits when those limits exist:
\begin{align}
\label{eq:sde_inference}
    \dd \Tilde{X}_t
    = & ~ [(1 -  \frac{\lambda_t}{2} \frac{\alpha_t}{\dot{\sigma}_t \sigma_t \alpha_t - \dot{\alpha}_t \sigma_t^2})
    u(\Tilde{X}_t,t)
    + \frac{\lambda_t}{2}
    \frac{\dot{\alpha_t}}{\dot{\sigma}_t \sigma_t \alpha_t - \dot{\alpha}_t \sigma_t^2} \Tilde{X}_t] \dd t + \sqrt{\lambda_t} \dd W_t.
\end{align}

\textbf{Markov Decision Processes.}
We consider a Markov Decision Process (MDP) with tuple $(\mathcal{S}, \mathcal{A}, P, R, \gamma)$.
Here, $\mathcal{S}$ and $\mathcal{A}$ denote the state and action spaces, 
$P(s' | s,a)$ is the transition kernel,
$R(s,a)$ is the reward function,
and $\gamma \in [0,1)$ is the discount factor.
A policy $\pi(a | s)$ induces a distribution over trajectories
$\tau = (s_0,a_0,s_1,a_1,\ldots)$, where $a_t \sim \pi(\cdot | s_t)$
and $s_{t+1} \sim P(\cdot | s_t,a_t)$.
The objective of RL is to maximize the expected cumulative return
\begin{align}
\label{eq:rl_objective_prelim}
J(\pi_\theta)
=
\E_{\tau \sim \pi_\theta}
\Big[
\sum_{t \ge 0} \gamma^t R(s_t,a_t)
\Big].
\end{align}
\textbf{Policy Optimization.}
Policy gradient methods (e.g., REINFORCE \cite{williams1992simple}) optimize the policy
$\pi_\theta(a|s)$ by approximating the gradient of the objective $J(\pi_\theta)$: 
\begin{align*}
\nabla_\theta J(\pi_\theta)
= & ~ 
\E_{\tau \sim \pi_\theta}
\Bigg[
\sum_{t \geq 0}
\gamma^t\nabla_\theta \log \pi_\theta(a_t| s_t)\,
r_t(s_t,a_t)
\Bigg], 
\quad \text{with} \quad
r_t(s_t,a_t) 
\coloneqq  
\sum_{\tau \geq t}
\gamma^{\tau-t} R(s_\tau,a_\tau),
\end{align*}
where $r_t(s_t,a_t)$ is the discounted cumulative future reward.
In practice, we replace $r_t(s_t,a_t)$ with an advantage estimator $A(s_t,a_t)$
to reduce the variance of gradient estimates.

\section{Flow Reinforcement Learning via Diffusion-Flow Duality}
\label{sec:framework}
\begin{figure*}[!htbp]
\centering
\includegraphics[width=\textwidth]{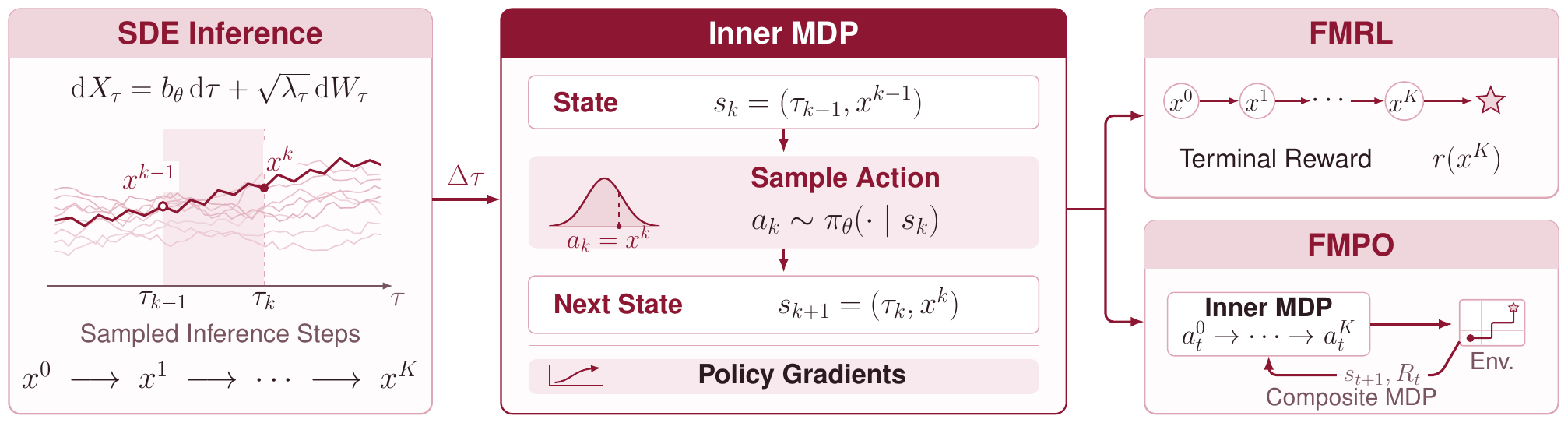}
\caption{SDE inference as an inner MDP for flow RL.
At step $k$, the state is $s_k=(\tau_{k-1},x^{k-1})$ and the action is the next sample $a_k=x^k$.
For positive-noise Euler--Maruyama steps, the inference kernel defines the Gaussian policy
$\pi_\theta(a_k\mid s_k)=p_{\tau_{k-1}}^\theta(x^k\mid x^{k-1})$.
FMRL attaches a terminal reward; FMPO composes the inner MDP with an environment MDP.}
\label{fig:flow_rl_framework}
\end{figure*}

We use SDE-based inference for flow matching
\eqref{eq:sde_inference} to construct a unified reinforcement learning
framework with two settings, illustrated in \cref{fig:flow_rl_framework}:
(i) flow matching RL fine-tuning (\cref{sec:fm_rl_finetune}) and
(ii) flow policy optimization (\cref{sec:fm_policy_op}).
The first treats stochastic flow inference as an \textit{inner} MDP for direct RL fine-tuning of a pretrained flow model
with terminal rewards.
The second composes this inner process with an
\textit{environment} MDP, yielding a \textit{composite} MDP for flow policy optimization.
Finally, our unified view recovers existing flow-based RL methods and yields new algorithms
(\cref{sec:unified_framework}).

\subsection{Flow Matching 
RL Fine-Tuning}
\label{sec:fm_rl_finetune}
Given a pretrained velocity field $u_\theta$,
let $r$ be a reward function and $p_\tau^\theta$ the probability
path induced by $u_\theta$.
\underline{F}low \underline{M}atching \underline{RL} (FMRL) maximizes expected terminal rewards,
\begin{align}
\label{eq:RL_objective}
    \mathcal{J}_{\text{FMRL}} = \E_{X_1 \sim p_1^\theta}[r(X_1)].
\end{align}
The SDE inference procedure in \eqref{eq:sde_inference} injects randomness into the inference process,
which allows us to reframe flow matching inference as an \textit{inner} MDP.
Use flow times $\tau_k=k/K$, $k=0,\ldots,K$, and step size $\Delta\tau=1/K$.
Write $x^k=x_{\tau_k}$ for the sample at inference step $k$.
Inspired by denoising diffusion policy optimization \cite{fan2023optimizing,black2023training},
we construct the following finite-horizon MDP:
\begin{align}
\label{eq:inner_mdp}
    s_k = & ~(\tau_{k-1},x^{k-1}), \quad
    \pi_\theta(a_k|s_k)=p_{\tau_{k-1}}^\theta(x^k|x^{k-1}), \nonumber\\
    a_k = & ~x^k, \quad
    P(\cdot|s_k,a_k)=\delta_{(\tau_k,a_k)}, \nonumber\\
    P_0 = & ~\delta_0\otimes\mathcal{N}(0,I), \quad
    R(s_k,a_k)=\begin{cases}
        r(x^K), & k=K,\\
        0, & 1\leq k<K.
    \end{cases}
\end{align}
where $k=1,\ldots,K$, the state $(1,x^K)$ is terminal, and $\delta_y$ denotes the Dirac delta distribution.
For conditional generation, include the fixed condition $c$ in the state, $s_k=(\tau_{k-1},x^{k-1},c)$, and condition all inference kernels on $c$ throughout the rollout. The finite-horizon inner return uses discount $1$.
The inner MDP has the same RL objective \eqref{eq:RL_objective}:
\begin{align*}
    \mathcal{J}(\theta)
    = \E_{a_k \sim \pi_\theta, s_k \sim P}[\sum_{k=1}^{K}
    R(s_k,a_k)]
    = \mathcal{J}_{\text{FMRL}}(\theta).
\end{align*}
We therefore optimize $\mathcal{J}(\theta)$ using standard
policy gradient methods.
For example,  Proximal Policy Optimization (PPO) \cite{schulman2017proximal} maximizes the surrogate objective
\begin{align*}
    \mathcal{J}_{\text{PPO}}(\theta)
    = \E_{(a_k,s_k) \sim \pi_{\text{old}}}
    [\min\{r_k(\theta) \widehat A_k,
    \Tilde{r}_k(\theta) \widehat A_k\}],
\end{align*}
where $\widehat A_k$ is an advantage estimate computed from the old-policy rollout and held fixed when differentiating with respect to $\theta$, and
\begin{align*}
    r_k(\theta) =\frac{\pi_\theta(a_k|s_k)}{\pi_{\text{old}}(a_k|s_k)}
    , ~
    \Tilde{r}_k(\theta)
    = \text{Clip}
    \left(r_k(\theta), 1-\epsilon,1+\epsilon \right).
\end{align*}
Here $p_{\tau_{k-1}}^\theta$ is the Euler--Maruyama kernel for step $k$.
Writing $b_\theta(x,\tau)$ for the drift in \eqref{eq:sde_inference} with $u$ replaced by $u_\theta$, a positive-noise step has
\begin{align*}
    p_{\tau_{k-1}}^\theta(x'|x)
    =\mathcal{N}\!\left(x';x+\Delta\tau\,b_\theta(x,\tau_{k-1}),
    \Delta\tau\,\lambda_{\tau_{k-1}}I\right).
\end{align*}
Thus the discretized transition likelihoods and their gradients are tractable.

\subsection{Flow Matching Policy Optimization}
\label{sec:fm_policy_op}
We consider an \textit{environment} MDP $\mathcal{M}_{\text{env}}=(\mathcal{S},\mathcal{A},P,R,\gamma)$.
At each environment step $t$, the agent observes $s_t \in \mathcal{S}$ and selects an action
$a_t \in \mathcal{A}$ according to a \textit{flow policy} $\pi_\theta(\cdot|s_t)$.
The objective of \underline{F}low \underline{M}atching \underline{P}olicy \underline{O}ptimization (FMPO) is to maximize the expected return
\begin{align*}
\mathcal{J}_{\text{env}}(\theta)
= \E_{a_t \sim \pi_\theta, s_t \sim P}[\sum_{t\geq0} \gamma^t R(s_t,a_t)].
\end{align*}
As shown in \cref{sec:fm_rl_finetune}, we view the SDE inference procedure \eqref{eq:sde_inference} of a flow model
as an \textit{inner} MDP.
Following diffusion policy policy optimization \cite{ren2024diffusion}, we combine the \textit{inner} MDP and \textit{environment} MDP into a \textit{composite} MDP.
This allows us to use policy gradient methods to optimize flow policy.

Use the same flow times $\tau_k=k/K$ and step size $\Delta\tau=1/K$ as in \cref{sec:fm_rl_finetune}.
At environment step $t$, SDE inference \eqref{eq:sde_inference} starts from $a_t^0\sim\mathcal{N}(0,I)$ and generates $a_t^k$ at flow time $\tau_k$, for $k=1,\ldots,K$.
We execute $a_t:=a_t^K$ in the environment.
$t$ indexes environment steps and $k$ inference steps.
We define the composite MDP $\mathcal{M}_{\text{comp}}$ as:
\begin{align}
\label{eq:composite_mdp}
    & ~ \bar{s}_{t,k} := (s_t,\tau_{k-1},a_t^{k-1}),
    \quad
    \bar{\pi}_\theta
    (\bar{a}_{t,k}|\bar{s}_{t,k}) := p_{\tau_{k-1}}^\theta(a_t^k|a_t^{k-1},s_t), \nonumber\\
    & ~ \bar{R}(\bar{s}_{t,k}, \bar{a}_{t,k}) 
    := 
    \begin{cases}
        0, & 1 \leq k < K,\\
        R(s_t, a_t^K), & 
        k = K,
    \end{cases} \quad
    \bar{a}_{t,k} := a_t^{k}, \nonumber\\
    & ~ \bar{s}'=
    \begin{cases}
        (s_t,\tau_k,a_t^k), & 1\leq k<K, \\
        (s_{t+1},\tau_0,a_{t+1}^0), & k=K.
    \end{cases}
\end{align}
Here $\bar{s}'$ is the next composite state under $\bar{P}$.
The kernel $p_{\tau_{k-1}}^\theta$ is the Euler--Maruyama step from \cref{sec:fm_rl_finetune}, conditioned on the environment state $s_t$.
At $k=K$, draw $s_{t+1}\sim P(\cdot|s_t,a_t^K)$ and fresh noise $a_{t+1}^0\sim\mathcal{N}(0,I)$ independently.
Inner transitions use discount $1$ for $k<K$, and discount $\gamma$ only at the environment boundary $k=K$.
Optimizing the expected return in the composite MDP is equivalent to
optimizing the original environment objective:
\begin{align*}
\mathcal{J}_{\text{com}}(\theta)
=
\E_{\bar{a}_{t,k} \sim \bar{\pi}_\theta,
\bar{s}_{t,k} \sim \bar{P}}
\Big[
\sum_{t,k} \gamma^t \bar{R}(\bar{s}_{t,k}, \bar{a}_{t,k})
\Big]
=
\mathcal{J}_{\text{env}}(\theta).
\end{align*}
For positive-noise Euler--Maruyama steps, the composite policy $\bar{\pi}_\theta$ admits a Gaussian likelihood, so standard policy gradient methods apply.

\subsection{Unified Framework and Instantiations}
\label{sec:unified_framework}

Our framework admits three algorithmic degrees of freedom:
(i) the choice of surrogate objective $\mathcal{J}(\theta)$;
(ii) the window over which stochastic differential equation (SDE) inference is applied;
and (iii) the noise schedule $\lambda_t$ used in SDE inference.
Different choices induce a broad family of flow-based RL algorithms, encompassing both existing methods and the instantiations proposed in this work.

\textbf{Flow Matching Policy Optimization with ChainPPO.}
For FM policy optimization in an environment MDP,
we instantiate the framework in \cref{sec:fm_policy_op} with a chain-level PPO surrogate,
yielding \textit{Flow Matching Policy Optimization with ChainPPO} (FMPO-CPPO).
Following \cref{sec:fm_policy_op}, the flow policy generates an environment action
$a_t=a_t^K$ through the stochastic inference chain
$a_t^0 \to a_t^1 \to \cdots \to a_t^K$.
The composite MDP \eqref{eq:composite_mdp} suggests treating this full inference chain as the stochastic action at environment step $t$.
Given samples collected from an old policy $\bar{\pi}_{\rm old}$, define
\begin{align*}
    r_{t,k}(\theta)
    =
    \frac{
    \bar{\pi}_\theta(\bar{a}_{t,k}|\bar{s}_{t,k})
    }{
    \bar{\pi}_{\rm old}(\bar{a}_{t,k}|\bar{s}_{t,k})
    },
    \qquad
    \rho_t^{\rm chain}(\theta)
    =
    \prod_{k=1}^K r_{t,k}(\theta).
\end{align*}
Since each inner action $a_t^k\in\mathbb{R}^{D}$ is an action chunk of dimension $D=Hd_a$ (horizon $H$, action dimension $d_a$), we normalize log-ratios by $D$ to make clipping ranges comparable across tasks:
\begin{align*}
    \tilde r_{t,k}(\theta)
    =
    r_{t,k}(\theta)^{1/D},
    \qquad
    \tilde\rho_t^{\rm chain}(\theta)
    =
    \prod_{k=1}^K \tilde r_{t,k}(\theta)
    =
    \exp\Big(\frac{1}{D}\sum_{k=1}^K \log r_{t,k}(\theta)\Big).
\end{align*}
FMPO-CPPO uses these tempered ratios in the chain-level PPO surrogate:
\begin{align*}
    \mathcal{J}_{\rm chain}(\theta)
    =
    \E_{\bar{\tau}_t \sim \bar{\pi}_{\rm old}}
    \left[
    \min\left\{
    \tilde\rho_t^{\rm chain}(\theta)\widehat A_t,\,
    \operatorname{Clip}\left(\tilde\rho_t^{\rm chain}(\theta),1-\epsilon_{\rm c},1+\epsilon_{\rm c}\right)\widehat A_t
    \right\}
    \right],
\end{align*}
where $\bar{\tau}_t=\{(\bar{s}_{t,k},\bar{a}_{t,k})\}_{k=1}^K$ and $\widehat A_t$ is the environment-level advantage estimate for the executed action $a_t=a_t^K$, held fixed during actor updates.
For the original chain ratio $\rho_t^{\rm chain}$, the clipping thresholds are $[(1-\epsilon_{\rm c})^D,(1+\epsilon_{\rm c})^D]$.
In practice, we optimize the mixed surrogate
\begin{align*}
    \mathcal{J}_{\rm FMPO\mbox{-}CPPO}(\theta)
    =
    \mathcal{J}_{\rm chain}(\theta)
    +
    \alpha \mathcal{J}_{\rm step}(\theta),
\end{align*}
where $\alpha \geq 0$ and $\mathcal{J}_{\rm step}$ is the standard PPO surrogate on the composite MDP \eqref{eq:composite_mdp}:
\begin{align*}
    \mathcal{J}_{\rm step}(\theta)
    =
    \E_{(\bar{s}_{t,k},\bar{a}_{t,k}) \sim \bar{\pi}_{\rm old}}
    \left[
    \min\left\{
    \tilde r_{t,k}(\theta)\widehat A_t,\,
    \operatorname{Clip}\left(\tilde r_{t,k}(\theta),1-\epsilon_{\rm s},1+\epsilon_{\rm s}\right)\widehat A_t
    \right\}
    \right].
\end{align*}
Before clipping, both the step-wise and chain-level surrogates have policy gradients at $\theta=\theta_{\rm old}$ proportional to that of the composite-MDP objective $\mathcal{J}_{\rm com}$ (\cref{sec:unbias_cppo}), which equals $\mathcal{J}_{\rm env}$ by \cref{sec:fm_policy_op}; the normalization only rescales them by $1/D$, since $\nabla_\theta\tilde\rho_t^{\rm chain}=\frac{1}{D}\nabla_\theta\log\rho_t^{\rm chain}$ at $\theta=\theta_{\rm old}$.
The step-wise term clips individual inference-transition surrogate contributions, while the chain term clips the full-chain surrogate contribution.
The chain-level term accounts jointly for all inference transitions producing the executed action, whereas the step-wise auxiliary term provides local surrogate contributions and improves gradient stability.
For positive-noise Euler--Maruyama steps, Gaussian transition kernels make these likelihood ratios tractable; see \cref{algo:FMPO_chainppo}.

\textbf{Unified Framework.}
Several recent flow-based RL methods instantiate our unified framework.
Flow-GRPO \cite{liu2025flow} instantiates flow matching RL fine-tuning (FMRL) with a GRPO surrogate.
DanceGRPO \cite{xue2025dancegrpo} follows the same design principles and fits FMRL.
Flow-SDE \cite{chen2025pi} instantiates FMPO using SDE inference and a PPO objective.
ReinFlow \cite{zhang2025reinflow} approximately instantiates FMPO with a PPO surrogate objective.
Unlike standard SDE inference in FM
\eqref{eq:sde_inference}, ReinFlow injects stochasticity through learnable noise.
Methods such as \cite{li2025mixgrpo, he2025tempflow} instantiate FMRL with restricted SDE inference windows to accelerate training.
We compare these methods with our FMPO-CPPO
in \cref{tab:unified_framework}.

\section{Theoretical Analysis}
\label{sec:theoretical_analysis}

We theoretically analyze the methods in
\cref{sec:framework}.
We first establish a correctness guarantee in \cref{sec:correctness}.
Under a small flow-matching error, the SDE-based training dynamics and the ODE-based deployment dynamics induce close terminal distributions.
We then study RL implications in \cref{sec:improvement_trans}.
A one-step SDE-based RL improvement transfers to the ODE-based deployment policy when flow-matching error is small.
Together, these results justify flow RL via diffusion--flow duality algorithms at both distributional and RL levels.
We consider the linear path $X_t= t X_1 + (1-t) X_0$ with independent $X_0\sim\mathcal{N}(0,I_d)$ and $X_1\sim p_1$.
For an early stopping time $0<T<1$, define
\begin{align*}
    \mathcal{L}_{\mathrm{FM},T}(\theta)
    &:=\int_0^T \E_{X_t\sim p_t^\star}
    [\|u_\theta(X_t,t)-u(X_t,t)\|_2^2]\,\dd t, \\
    \epsilon_\theta&:=\sqrt{\mathcal{L}_{\mathrm{FM},T}(\theta)}.
\end{align*}
Here $p_t^\star$ is the reference path induced by the true velocity $u$.
The unindexed $\epsilon$ in an error-bound assumption denotes an upper bound on $\epsilon_\theta$.

\subsection{Correctness}
\label{sec:correctness}

For an estimated velocity field $u_\theta$, the identity 
\eqref{eq:velocity_score} does not always hold.
Consequently, the ODE inference dynamics in \eqref{eq:ode_process} and the SDE inference dynamics in \eqref{eq:sde_process} are generally not equivalent.
This ODE--SDE mismatch creates two inconsistencies in the algorithms in \cref{sec:framework}.

First, consider a pretrained velocity field $u_{\theta_1}$ before
reinforcement learning.
At this stage, the base policy follows ODE inference, while
exploration during RL relies on SDE inference.
Second, consider the velocity field $u_{\theta_2}$ obtained after
reinforcement learning.
Although $u_{\theta_2}$ is optimized under SDE dynamics during training,
deployment again follows ODE-based inference for efficiency.
In both cases, the transition between ODE and SDE inference is not
exactly equivalent, since the parameterized velocities $u_{\theta_1}$ and $u_{\theta_2}$ differ from the true affine-path-induced velocity $u$.

Fortunately, when the estimated velocity $u_\theta$ stays close to the true velocity $u$, identity \eqref{eq:velocity_score} holds approximately.
The following theorem formalizes this observation.
\begin{theorem}
[Correctness for Flow RL via Stochastic Inference, Informal]
\label{thm:correctness}
Fix an early stopping time $0<T<1$.
Let $p_T$ and $q_T$ be the terminal distributions of ODE inference \eqref{eq:ode_process} and SDE inference \eqref{eq:sde_inference}, respectively, using $u_\theta$ from the same Gaussian source.
Assume $p_1$ has a finite second moment and the velocities are continuous in time.
Suppose $u$ is continuously differentiable and globally Lipschitz in space, uniformly over $t\in[0,T]$.
Suppose $u_\theta$ is twice continuously differentiable in space, with uniformly bounded Jacobian and gradient of its divergence over $x,t,\theta$.
Assume the noise schedule $\lambda$ is continuous and strictly positive on $[0,T]$.
If $\epsilon_\theta\le\epsilon$, then
\begin{align*}
    \text{TV}(p_T||q_T)\leq O(\sqrt{\epsilon})
\end{align*}
as $\epsilon\to0$, with a constant independent of $\theta$ and $\epsilon$.
\end{theorem}

\begin{proof}
    Please see \cref{sec:proof_correctness} for the formal statement and detailed proof.
\end{proof}

\cref{thm:correctness} establishes correctness for our Flow Matching RL framework under a fine-tuning setting.
Before the first transition, we require pretraining.
This ensures that the base policy $u_{\theta_1}$ stays close to the true constructed velocity.
The base policy then satisfies condition \eqref{eq:similar_velocity}.
For the second transition, we assume the RL algorithm performs fine-tuning without large changes to the velocity.
Under this assumption, $u_{\theta_2}$ remains close to $u_{\theta_1}$ and to the true constructed velocity.
The fine-tuned policy then satisfies condition \eqref{eq:similar_velocity}.

\begin{figure*}[t]
\centering

\begin{subfigure}[c]{0.32\linewidth}
  \centering
  \includegraphics[width=\linewidth]{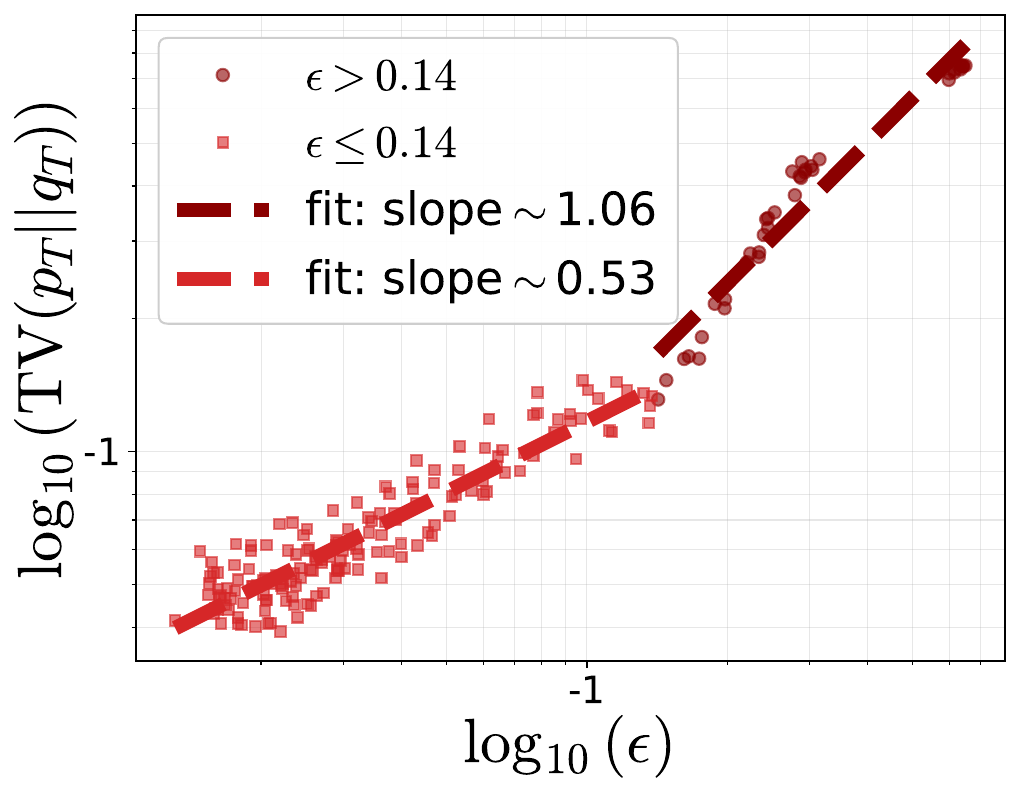}
  \centering
  \caption*{\small \textbf{(a) $\text{TV}(p_T||q_T)$ versus $\epsilon$ in log scale}}
\end{subfigure}\hfill\begin{subfigure}[c]{0.32\linewidth}
  \centering
  \includegraphics[width=\linewidth]{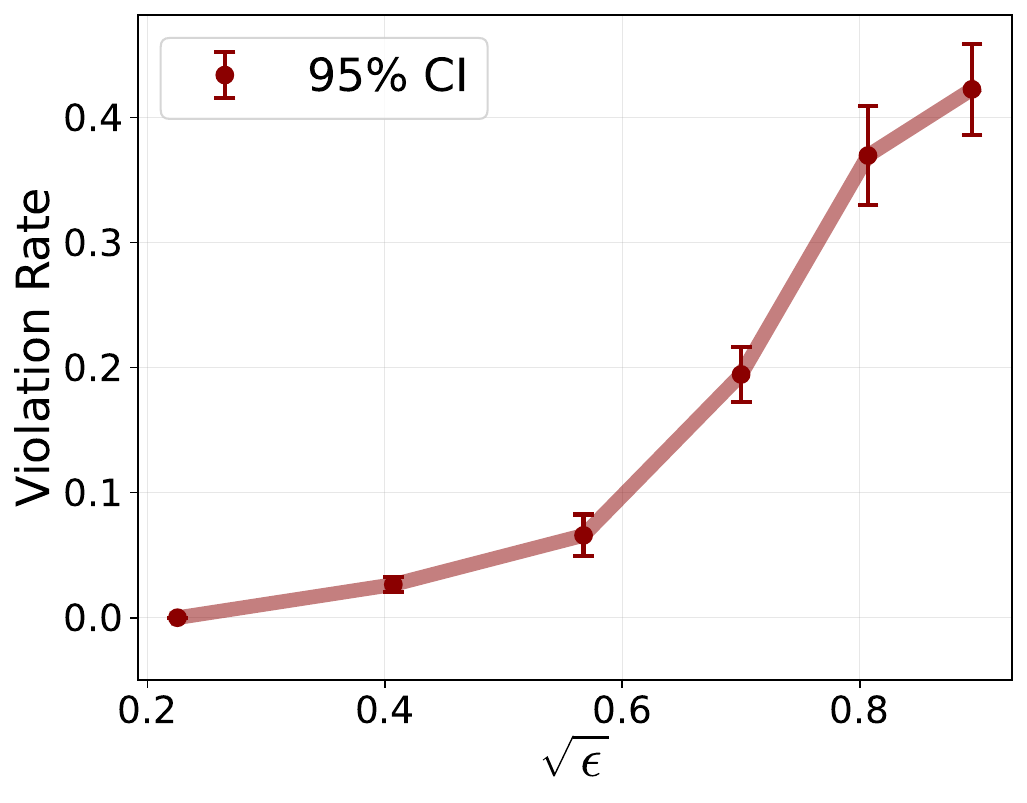}
  \centering
  \caption*{\small\textbf{(b) One-step improvement violation rate versus $\sqrt\epsilon$}}
\end{subfigure}\hfill\begin{subfigure}[c]{0.32\linewidth}
  \centering
  \includegraphics[width=\linewidth]{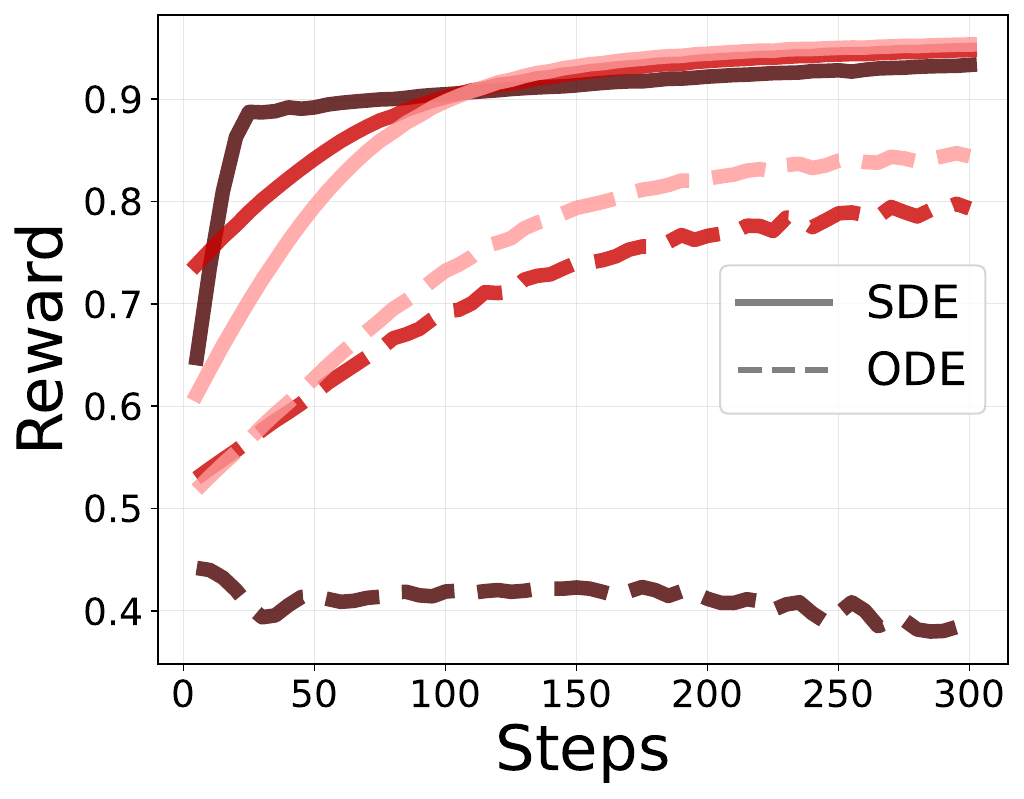}
  \centering
  \caption*{\small\textbf{(c) SDE and ODE reward trajectories
 }}
\end{subfigure}

\caption{\small \textbf{(a) $\mathrm{TV}(p_T \| q_T)$ versus $\epsilon$.} 
Recall the definition of flow-matching error as
$\epsilon^2 = \int_0^T \E_{X_t \sim p_t^\star}
[\|u(X_t,t)-u_\theta(X_t,t)\|_2^2] \dd t$.
For small errors ($\epsilon \le 0.14$), the fitted slope ($\approx 0.53$) matches the theoretical $O(\sqrt{\epsilon})$ rate.
For larger errors ($\epsilon > 0.14$), the fitted slope ($\approx 1.06$) corresponds to the $O(\epsilon)$ regime, consistent with the bound in \cref{thm:correctness}.
\textbf{(b) Violation rate versus $\sqrt\epsilon$.} 
The violation rate measures the fraction of consecutive checkpoint pairs where SDE reward improves but ODE reward decreases. 
Pairs are binned by $\sqrt{\epsilon}$. The rate increases with $\sqrt{\epsilon}$, confirming better SDE-to-ODE transfer for small $\epsilon$.
\textbf{(c) Training-time SDE and deployment-time ODE reward trajectories.}
Solid curves show rewards under training-time SDE inference; dashed curves show rewards under deployment-time ODE inference.
Lighter curves indicate better-pretrained base models with smaller $\epsilon$.
ODE rewards closely track SDE rewards for well-pretrained models, but oscillate for poorly pretrained models (darkest) despite steady SDE improvement, validating the failure mode in \cref{rem:limit}.}

\label{fig:three}
\vspace{-1em}
\end{figure*}

\subsection{SDE-to-ODE Performance Transfer}
\label{sec:improvement_trans}
\cref{sec:correctness} establishes conditions for distributional consistency between SDE- and ODE-based inference dynamics. Under the same assumptions, we extend this analysis to RL. The distributional correctness guarantee implies a one-step performance improvement bound for the ODE inference policy.
\begin{theorem}[Transfer of Improvement from SDE to ODE]
\label{thm:improve_trans}
Let $r: \R^d \rightarrow \R$ be the terminal reward, with $\abs{r(x)} < R < \infty$ for any $x \in \R^d$.
Fix early stopping time $0<T<1$ and noise level $\lambda_t >0$.
For each parameter $\theta$, we define SDE inference process $Z_t^\theta$ and ODE inference $Y_t^\theta$ as in \cref{thm:correctness_re}.
Define the corresponding performances
\begin{align*}
    J_{\text{SDE}}(\theta)
    := \E[r(Z^\theta_T)],
    \quad
    J_{\text{ODE}}(\theta)
    := \E[r(Y^\theta_T)].
\end{align*}
Let $\{\theta_k\}_{k\geq0}$ be the parameter sequence of an RL algorithm satisfying the following one-step SDE performance improvement condition:
\begin{align*}
    J_{\text{SDE}}(\theta_{k+1})
    \geq J_{\text{SDE}}(\theta_{k}) + \Delta_k, 
    \quad \Delta_k > 0.
\end{align*}
Assume that the hypotheses of \cref{thm:correctness_re} hold for every $\theta_k$ with the same constants $L,H$ and noise schedule.
Then the ODE performance satisfies 
\begin{align*}
    J_{\text{ODE}}(\theta_{k+1})
    \geq J_{\text{ODE}}(\theta_{k}) + \Delta_k - 2R(\delta_k + \delta_{k+1}),
\end{align*}
where $\delta_k := C_{\mathrm{ODE}}\sqrt{\epsilon_k}+\frac{\sqrt{C_{\lambda,T}}}{2}\epsilon_k=O(\sqrt{\epsilon_k})$ bounds the total variation distance
between the terminal distributions induced by the ODE and SDE dynamics at
iteration $k$, and $\epsilon_k := \sqrt{\mathcal{L}_{\mathrm{FM},T}(\theta_k)}$
denotes the FM error for $\theta_k$.
\end{theorem}

\begin{proof}
    Please see \cref{sec:proof_improve_trans} for a detailed proof.
\end{proof}

\begin{remark}[Practical Implication]
\label{rem:limit}
The performance transfer guarantee requires the estimated velocity $u_\theta(x,t)$ to stay close to the ground-truth affine-path velocity $u(x,t)$.
This condition holds in fine-tuning settings, where the pretrained flow model has small flow-matching error and RL updates remain moderate.
In contrast, RL from scratch produces velocities far from the affine-path dynamics, creating a large ODE-SDE mismatch.
Thus, our analysis identifies pretraining quality and controlled RL drift as key factors for reliable transfer from SDE-based training to ODE-based deployment.
\end{remark}

By \cref{thm:improve_trans}, the deployment-time ODE return is nondecreasing whenever $\Delta_k \geq 2R(\delta_k+\delta_{k+1})$.
Smaller flow-matching error reduces this threshold and the SDE improvement needed for transfer to ODE deployment.

\section{Experimental Studies}
\label{sec:exp}
In this section, \cref{sec:validate_corr} and \cref{sec:validate_trans} present numerical experiments to validate the correctness theorem (\cref{thm:correctness}) and improvement transfer theorem (\cref{thm:improve_trans}).
In addition, \cref{sec:algorithm_eva} performs experiments to demonstrate the effectiveness of our method FMPO-CPPO (\cref{algo:FMPO_chainppo}).

\subsection{Validating Correctness (\texorpdfstring{\cref{thm:correctness}}{Theorem~\ref{thm:correctness}})}
\label{sec:validate_corr}

\textbf{Setup.}
Let $\epsilon^2 =      \int_0^T \E_{
    X_t \sim p^\star_t}
    [\|u(X_t,t)-u_\theta(X_t,t)\|_2^2]
    \dd t $ be the flow matching loss.
Let $p_T$ and $q_T$ be the distributions induced by deployment-time ODE and training-time SDE inference, respectively (as in \cref{thm:correctness}).
We validate the bound in \cref{thm:correctness},
\begin{align*}
    \text{TV}(p_T||q_T)
    \leq O(\sqrt{\epsilon}).
\end{align*}

\textbf{Data.}
We consider a synthetic setting in $\mathbb{R}^2$ with $p_0 = N(0,I_2)$ and $p_1 = N(\mu,\Sigma)$. 
With a linear probability path, the induced velocity field admits the closed-form expression
\begin{align*}
    u^*(x, t) = \mu +(t\Sigma-(1-t)I)C_t^{-1}(x-t\mu),
\end{align*}
where $    \mu = 
\begin{pmatrix}
2 \\
-1
\end{pmatrix}
, ~
\Sigma =
\begin{bmatrix}
0.7 & 0.25 \\
0.25 & 0.5
\end{bmatrix},~
C_t=(1-t)^2I+t^2 \Sigma$.

\textbf{Results.}
In \cref{fig:three}(a), we numerically validate \cref{thm:correctness} by training a velocity network $u_\theta$ to approximate the closed-form Gaussian-to-Gaussian velocity field $u^*$.
We vary approximation quality through training iterations and compute the flow-matching error $\epsilon$. We observe two-regime scaling consistent with the theoretical bound.
Under the assumptions of \cref{thm:correctness}, \cref{sec:proof_correctness} gives
\begin{align*}
   \text{TV}(p_T||q_T)\leq C_{\mathrm{ODE}}\sqrt{\epsilon}+\frac{\sqrt{C_{\lambda,T}}}{2}\epsilon,
\end{align*}
where $C_{\mathrm{ODE}}$ and $C_{\lambda,T}$ are defined in \cref{sec:main_proof_correctness} and $T$ is the early stopping time.
For larger errors ($\epsilon > 0.14$), the linear $\epsilon$ term dominates, yielding $\text{TV}(p_T||q_T) = O(\epsilon)$.
The fitted log-log slope is approximately $1.06$, consistent with this prediction. For smaller errors ($\epsilon \leq 0.14$), the $C_{\mathrm{ODE}}\sqrt{\epsilon}$ term dominates, yielding $\text{TV} = O(\sqrt{\epsilon})$.
The fitted slope converges to $0.53$, matching the $O(\sqrt{\epsilon})$ rate predicted by \cref{thm:correctness}.
These results confirm that ODE--SDE TV distance scales as $O(\sqrt{\epsilon})$ in the fine-tuning regime where $\epsilon$ is small.

\subsection{Validating Performance Transfer (\texorpdfstring{\cref{thm:improve_trans}}{Theorem~\ref{thm:improve_trans}})}
\label{sec:validate_trans}
\textbf{Setup.} 
We validate \cref{thm:improve_trans} and \cref{rem:limit} by fine-tuning pretrained base models spanning different flow-matching error levels.
From each base model, we run flow matching RL fine-tuning via SDE inference and evaluate checkpoints under both training-time SDE and deployment-time ODE inference.
Checkpoints are saved every $5$ RL training steps to capture fine-grained reward transitions.

\textbf{Data.}
We work in $\mathbb{R}^2$ with $p_0 = N(0,I_2)$ and $p_1 = N(\mu,\Sigma)$. With a linear path, the velocity is
\begin{align*}
    u^*(x, t) = \mu +(t\Sigma-(1-t)I)C_t^{-1}(x-t\mu),
\end{align*}
where $    \mu = 
\begin{pmatrix}
2.5\\
2.5
\end{pmatrix}
,~
\Sigma =
\begin{bmatrix}
0.25 & 0.18 \\
0.18 & 0.80
\end{bmatrix},~
C_t=(1-t)^2I+t^2\Sigma$.
The terminal reward is bounded in $(0,1]$ and is defined as a temperature-scaled Mahalanobis score:
\begin{align*}
    r(x) = \exp{-1.5(x-\mu)^\top \Sigma^{-1}(x-\mu)}.
\end{align*}

\textbf{Metric.}
We use the violation rate to validate \cref{thm:improve_trans}.
For consecutive checkpoint pairs $(k,k+1)$, let $\Delta_{\rm SDE} = R^{(k+1)}_{\rm SDE} - R^{(k)}_{\rm SDE}$, $\Delta_{\rm ODE} = R^{(k+1)}_{\rm ODE} - R^{(k)}_{\rm ODE}$. 
The violation rate is:
\begin{align*}
\text{Violation Rate}
=
\frac{\sum_{k} \one\{\Delta_{\text{SDE}}>0,\ \Delta_{\text{ODE}}<0\}}
     {\sum_{k} \one\{\Delta_{\text{SDE}}>0\}} .
\end{align*}
Among pairs with SDE reward improvement (the one-step improvement condition of \cref{thm:improve_trans}), the violation rate is the fraction with ODE reward decline. We report the violation rate for checkpoint pairs binned by $\sqrt{\epsilon}$.

\begin{figure*}[t]
\centering

\begin{subfigure}[c]{0.24\linewidth}
  \centering
  \includegraphics[width=\linewidth]{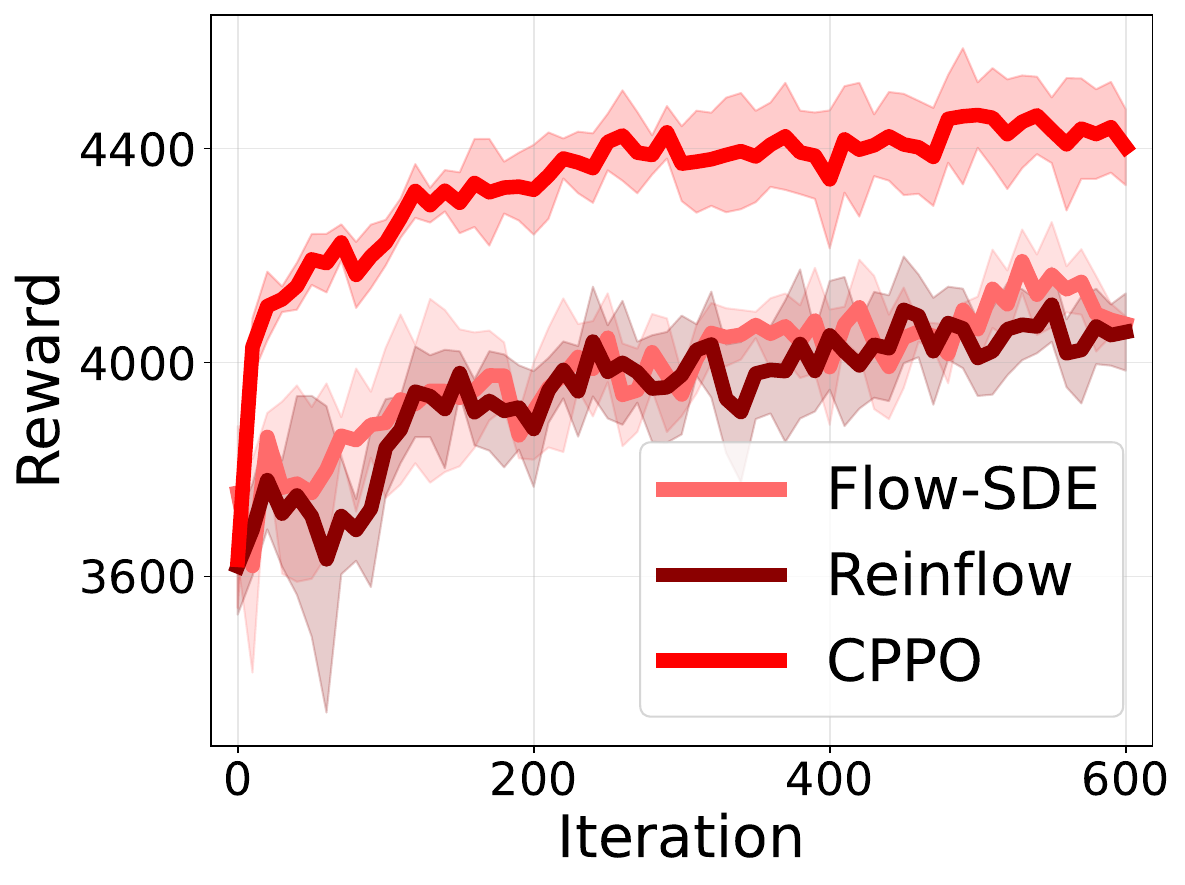}
  \caption*{\textbf{(a) Ant-v2}}
\end{subfigure}\hfill\begin{subfigure}[c]{0.24\linewidth}
  \centering
  \includegraphics[width=\linewidth]{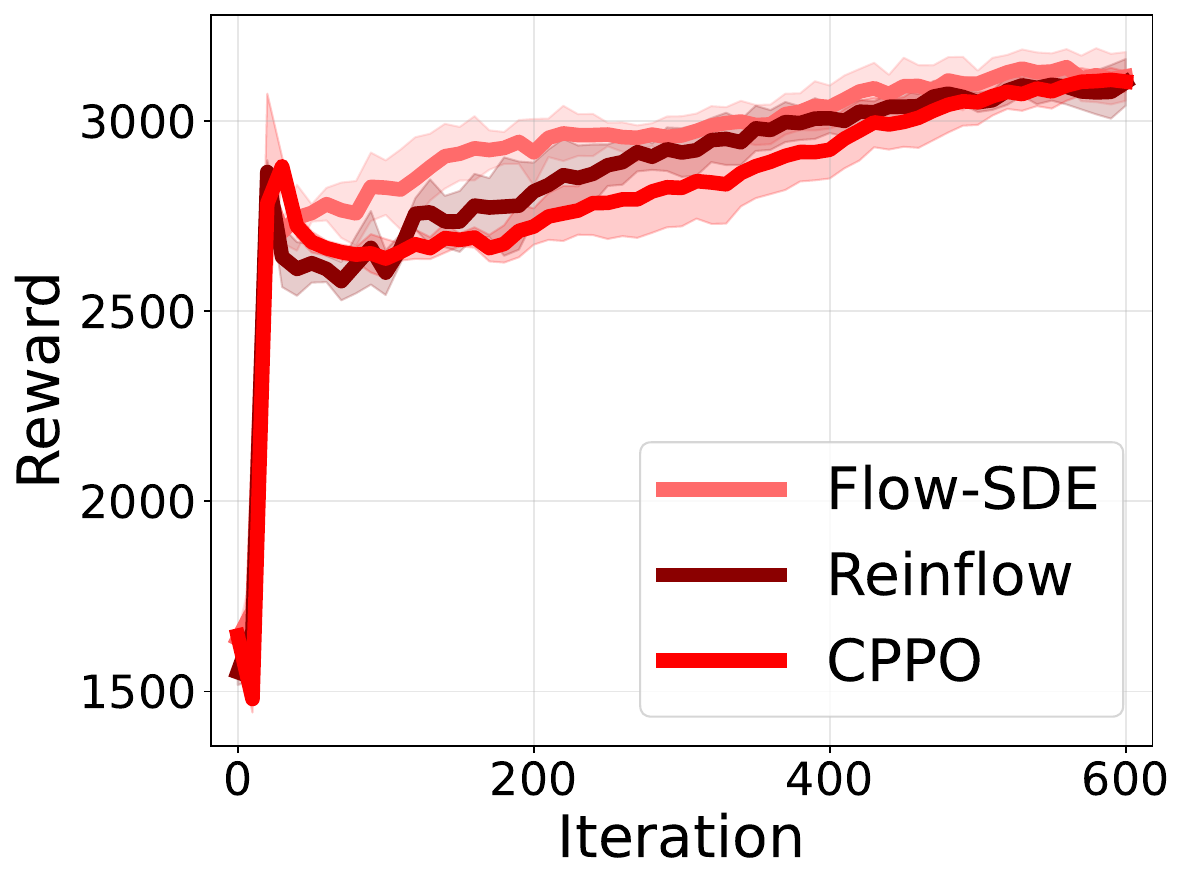}
  \caption*{\textbf{(b) Hopper-v2}}
\end{subfigure}\hfill\begin{subfigure}[c]{0.24\linewidth}
  \centering
  \includegraphics[width=\linewidth]{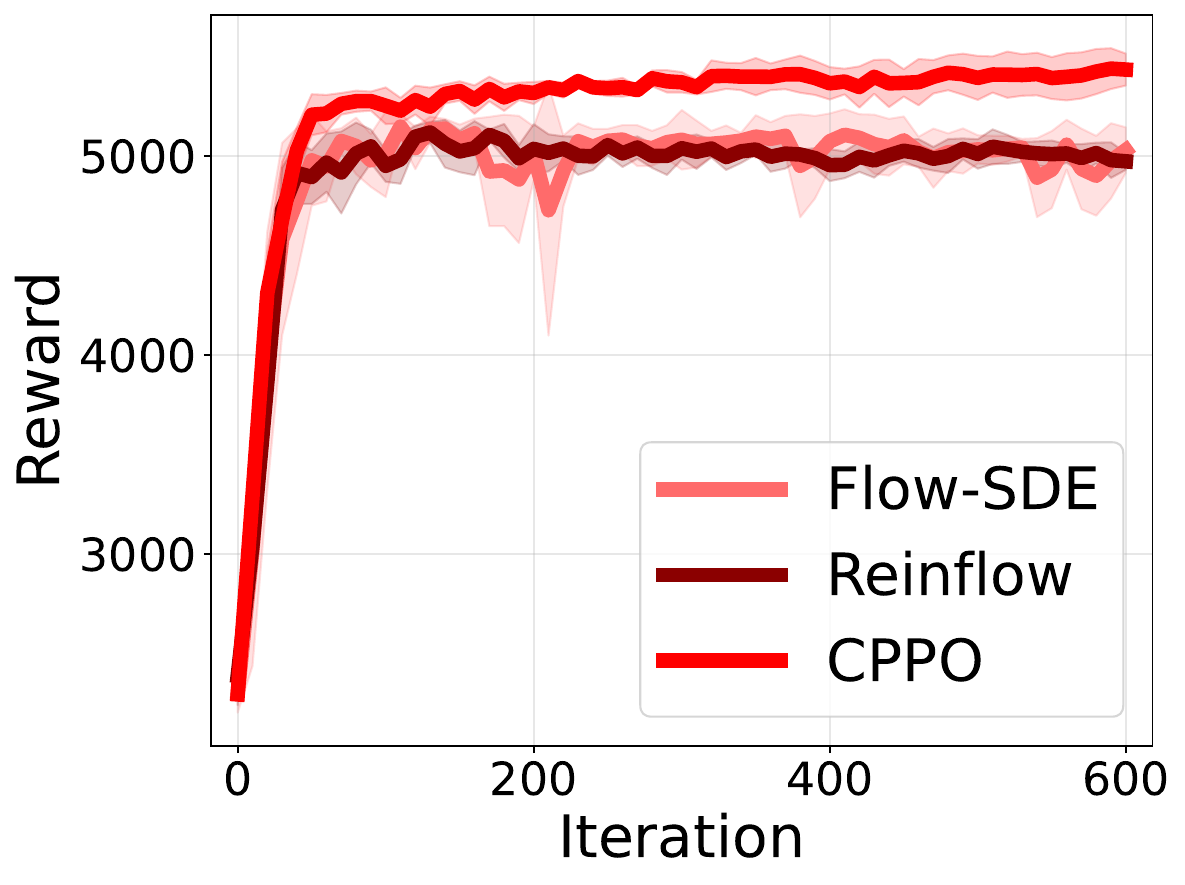}
  \caption*{\textbf{(c) Humanoid-v3}}
\end{subfigure}
\begin{subfigure}[c]{0.24\linewidth}
  \centering
  \includegraphics[width=\linewidth]{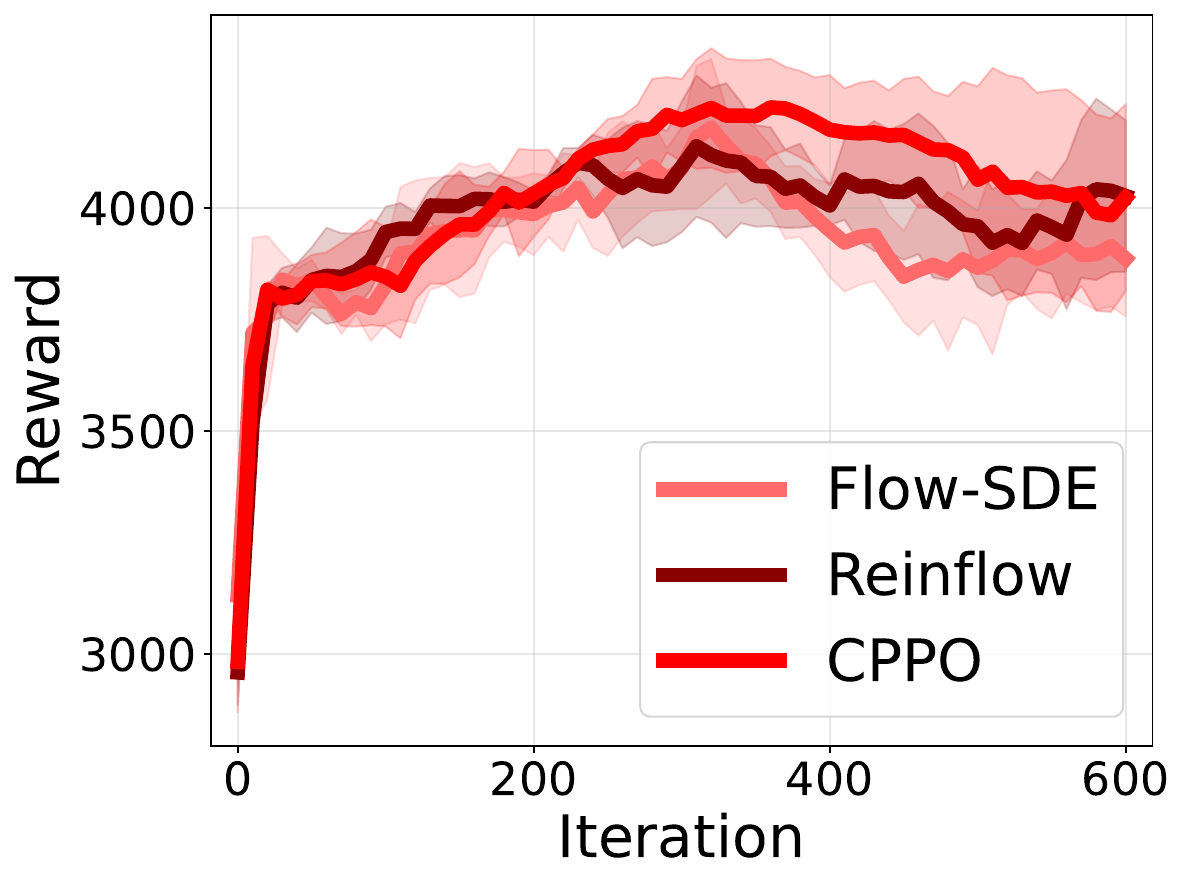}
  \caption*{\textbf{(d) Walker-v2}}
\end{subfigure}

\caption{\small Learning curves of evaluation reward on four OpenAI Gym MuJoCo environments: \textbf{(a) Ant-v2}, \textbf{(b) Hopper-v2}, \textbf{(c) Humanoid-v3}, and \textbf{(d) Walker-v2}. Solid lines show mean performance over 5 random seeds; shading indicates one standard deviation. FMPO-CPPO achieves higher reward than ReinFlow and Flow-SDE on Ant-v2, Humanoid-v3 and Walker-v2 and comparable performance on Hopper-v2.}
\label{fig:FMPO_exp}
\vspace{-1.5em}
\end{figure*}

\textbf{Results.}
The inequality suggests that SDE-to-ODE transfer is more reliable when the flow-matching error $\epsilon$ is small, since smaller $\epsilon$ reduces ODE–SDE mismatch and the penalty term $R(\delta_k+\delta_{k+1})$. We therefore measure the ODE monotonicity violation rate conditioned on one-step SDE improvement, $\Pr(\Delta_{\rm ODE}<0\mid \Delta_{\rm SDE}>0)$, and plot it against $\epsilon$. Empirically, this violation probability increases with $\epsilon$, supporting the insight from \cref{thm:improve_trans}: SDE fine-tuning transfers more reliably to ODE for low-$\epsilon$ base models, while high-$\epsilon$ bases show weaker transfer and more frequent violations.
We also observe the failure mode predicted by \cref{rem:limit} in \cref{fig:three}(c). For large-$\epsilon$ bases, the SDE reward improves during RL fine-tuning, but the ODE reward only oscillates around its initial value. In contrast, small-$\epsilon$ bases show effective SDE-to-ODE transfer. This confirms that the transfer guarantee requires small flow-matching error, a condition typical of fine-tuning but not training from scratch.

\subsection{FMPO-CPPO Evaluation}
\textbf{Setup.}
We evaluate FMPO-CPPO, ReinFlow~\cite{zhang2025reinflow} and Flow-SDE~\cite{chen2025pi} on four continuous-control locomotion benchmarks from OpenAI Gym~\cite{brockman2016openai}:
Hopper, Walker2d, Ant, and
Humanoid, ordered by increasing task difficulty. Each task is averaged over 5 random seeds. We put ablations in \cref{sec:abl}.

\textbf{Results.}
As shown in \cref{fig:FMPO_exp}, FMPO-CPPO attains higher evaluation reward on Ant-v2, Walker-v2 and Humanoid-v3, while showing comparable performance on Hopper-v2. The gains are most pronounced on the higher-dimensional Humanoid-v3 and Ant-v2 tasks, where FMPO-CPPO converges to a stronger final policy and improves reward faster early in training.
\label{sec:algorithm_eva}

\subsection{Validating Correctness and Performance Transfer in the Image Generation Setting}

\begin{figure*}[t]
    \centering

    \begin{subfigure}[c]{0.24\linewidth}
      \centering
      \includegraphics[width=\linewidth]{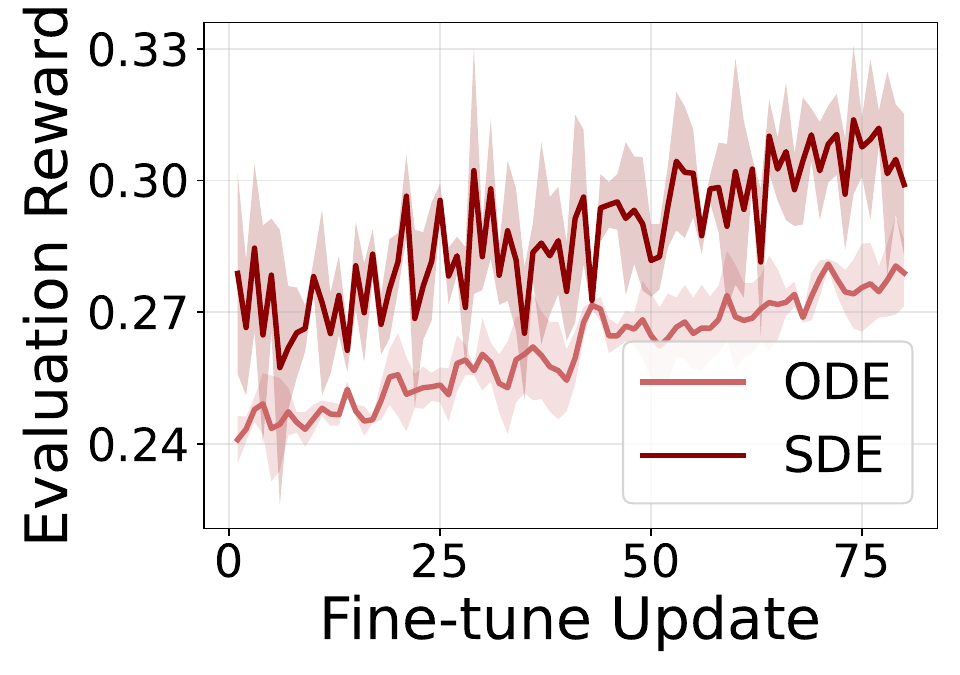}
      \caption*{\small \textbf{(a) Weak}}
    \end{subfigure}\hfill
    \begin{subfigure}[c]{0.24\linewidth}
      \centering
      \includegraphics[width=\linewidth]{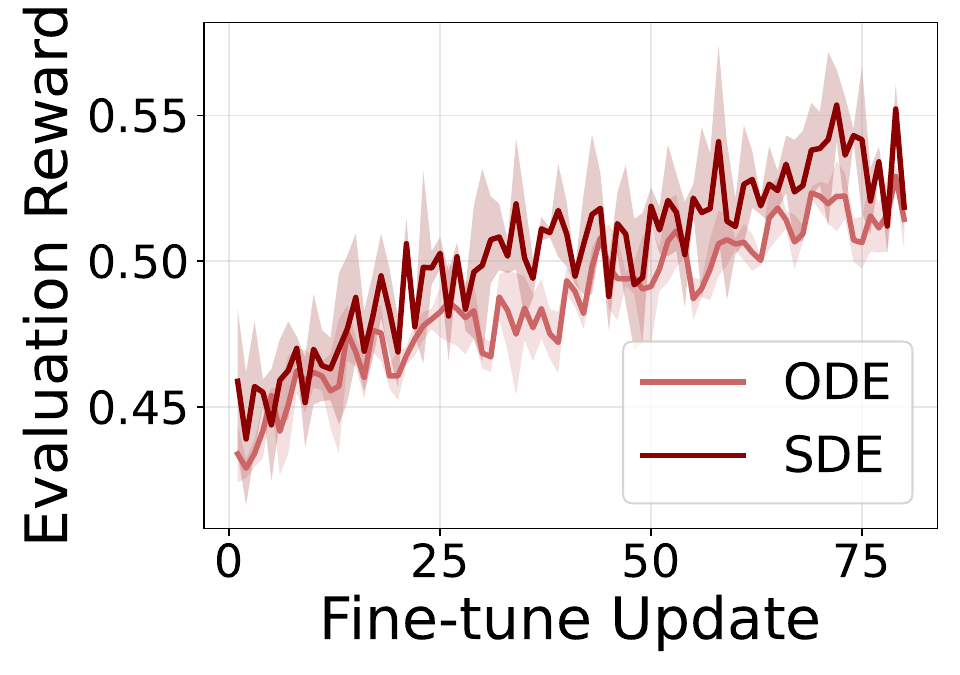}
      \caption*{\small \textbf{(b) Medium}}
    \end{subfigure}
    \begin{subfigure}[c]{0.24\linewidth}
      \centering
      \includegraphics[width=\linewidth]{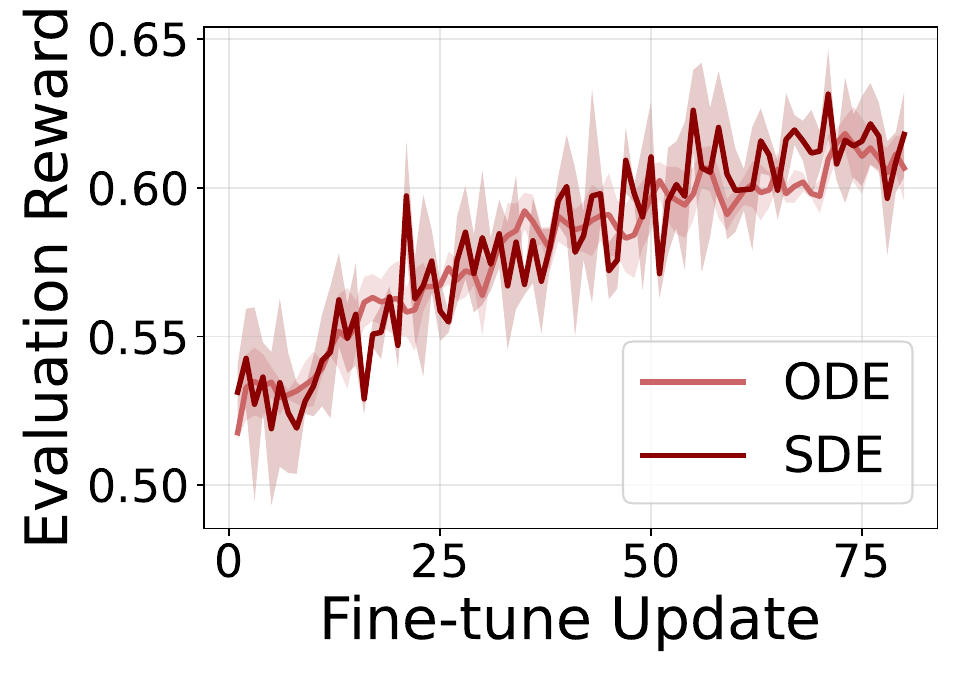}
      \caption*{\small \textbf{(c) Strong}}
    \end{subfigure}\hfill
    \begin{subfigure}[c]{0.24\linewidth}
      \centering
      \includegraphics[width=\linewidth]{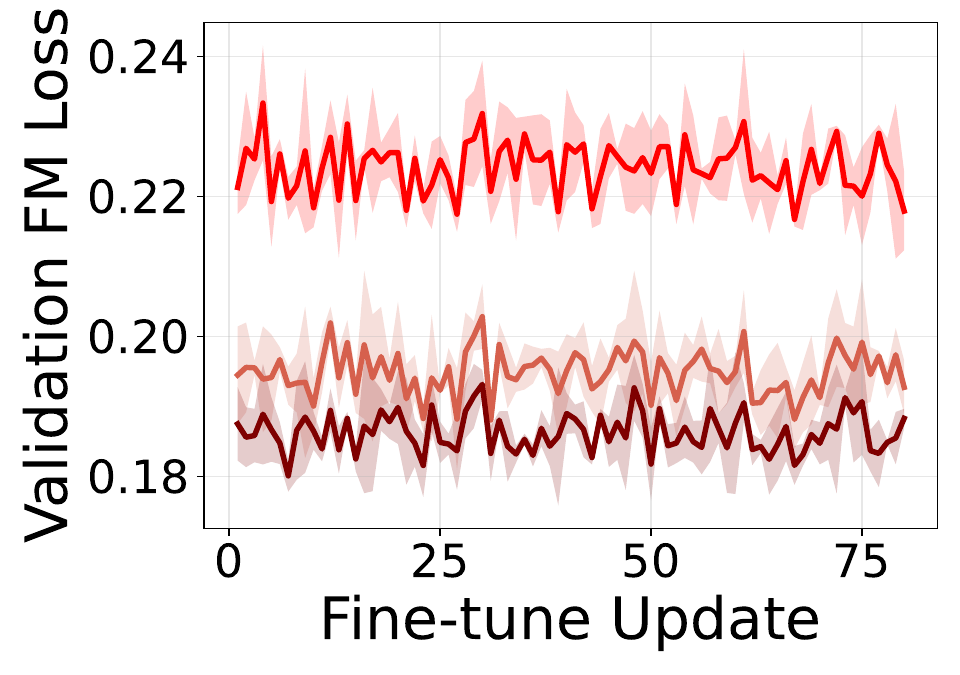}
      \caption*{\small \textbf{(d) Flow matching error}}
    \end{subfigure}
    \vspace{-.5em}
    \caption{\small \textbf{(a) Weak, (b) Medium, and (c) Strong:} ODE and SDE fine-tuning reward trajectories of the three pretrained base models. \textbf{(d) FM error:} Validation flow-matching error of these models during RL fine-tuning. The dark red, orange, and red lines represent Strong, Medium, and Weak, respectively. Curves show means over three independent seeds, and shaded bands show one standard deviation.}
    \vspace{-1.5em}
    \label{fig:FMRL_val}
\end{figure*}

\textbf{Setup.}
We validate \cref{thm:correctness} and \cref{thm:improve_trans} on class-conditional CIFAR-10 generative fine-tuning. We use validation-set conditional flow matching loss as a substitute for the flow-matching error since the ground-truth velocity is unobtainable. We train three base models: \textbf{Weak}, \textbf{Medium}, \textbf{Strong} with validation flow-matching error of $0.225$, $0.196$, $0.184$, respectively. We fine-tune the three base models with Flow-GRPO and evaluate ODE and SDE rewards after each update.

\textbf{Results.}
The reward trajectories in \cref{fig:FMRL_val} exhibit the trend predicted by \cref{thm:correctness}. The ODE-SDE performance gap decreases as pretraining quality improves. The trajectories show visible separation for the Weak model and substantial overlap for the Strong model. We compute the violation rate defined in \cref{sec:validate_trans} on seed-averaged trajectories. The violation rates are $0.467$, $0.419$, and $0.413$ for Weak, Medium, and Strong, respectively, exhibiting the trend predicted by \cref{thm:improve_trans}. Validation flow matching errors also fluctuate within a bounded range throughout fine-tuning (\cref{fig:FMRL_val}(d)). This is consistent with the fine-tuning regime of our theories: SDE-based RL updates improve the reward without meaningfully drifting the underlying velocity field away from the pretrained one.

\section{Discussion and Conclusion}
\label{sec:conclusion}
We introduce a unified framework using SDE inference for flow matching RL fine-tuning (\cref{sec:fm_rl_finetune}) and flow matching policy optimization (\cref{sec:fm_policy_op}).
We introduce \textsc{FMPO-CPPO} (\cref{algo:FMPO_chainppo}) as an FMPO instantiation and show that the framework subsumes several recent flow-based RL methods (\cref{sec:unified_framework}), summarized in \cref{tab:unified_framework}.

Theoretically, we establish two key guarantees.
First, we prove a correctness result: when the flow-matching error is small, training-time SDE and deployment-time ODE inference induce close terminal distributions (\cref{thm:correctness}).
Second, we derive a performance transfer guarantee: SDE-based policy improvements transfer to ODE-based deployment up to an error term controlled by the flow-matching error (\cref{thm:improve_trans}).

We numerically validate both theoretical predictions (\cref{sec:validate_corr} and \cref{sec:validate_trans}).
We also evaluate \textsc{FMPO-CPPO}, which achieves performance comparable to or better than existing baselines across representative continuous-control tasks (\cref{sec:algorithm_eva}).

Future work will study RL effectiveness beyond fine-tuning, including training flow policies from scratch.

\section*{LLM Usage Disclosure}
We used large language models (LLMs) to aid and polish writing, such as improving clarity, grammar, and conciseness. We also used LLMs for retrieval and discovery, for example searching the literature to identify potential missing related work. All technical content, proofs, experiments, and results are original contributions by the authors.

\clearpage
\appendix
\startcontents[appendices]
\part*{Appendix}
\begingroup
\setlength{\parskip}{0pt}
\printcontents[appendices]{}{1}{}
\endgroup

\section{Related Works}
\label{sec:related_works}

\textbf{Flow Matching and Flow Matching Policy.}
Flow matching 
\cite{lipman2022flow, albergo2023stochastic, liu2022flow} emerges as a powerful class of generative modeling techniques, achieving strong empirical performance across a wide range of domains, including image, speech, and video generation \cite{esser2024scaling, le2023voicebox, polyak2025moviegencastmedia}.
Recent works extend flow matching to policy learning \cite{black2024pi}, where the generative model represents a control policy or action distribution.
Flow-based vision-language-action models demonstrate state-of-the-art performance on complex manipulation and control tasks \cite{intelligence2025pi5,intelligence2025pi}, highlighting the potential of flow-based policies.

\textbf{Flow Matching Error Bounds.}
A line of recent work develops theoretical error bounds for fully deterministic flow matching models under different notions of distributional discrepancy.
Early analyses focus on Wasserstein distances, where \citet{albergo2022building} establish a 2-Wasserstein bound assuming uniform Lipschitz regularity of the velocity field, and \citet{benton2023error} later relax this requirement to time-dependent regularity conditions.
Beyond Wasserstein metrics, \citet{li2023towards} derive a total variation bound, but their result relies on the estimated score error being small.
More recently, \citet{su2025flow} provide a non-asymptotic KL divergence bound that depends only on the regularity and smoothness of the data path and velocity fields, without imposing additional regularity assumptions on the estimated score path.

\textbf{RL Fine-Tuning for Flow-Based Policies.}
Aligned with our main text, we consider two closely related settings: 
flow matching RL fine-tuning, where no external environment MDP is involved (e.g., image generation), and 
flow matching policy optimization, where the flow-based policy interacts with an environment MDP (e.g., robotics).

For flow matching RL fine-tuning, early work such as \citet{fan2025online} adopts reward-weighted regression methods to fine-tune flow matching models with Wasserstein regularization.
More recent approaches~\cite{liu2025flow, xue2025dancegrpo, li2025mixgrpo, he2025tempflow} introduce stochastic differential equation (SDE) inference during training and apply policy gradient methods with GRPO surrogate objectives.
These methods are special instantiations of our flow matching RL fine-tuning framework under particular choices of surrogate objectives, inference windows, and noise schedules.

For flow matching policy optimization, which involves an external environment MDP, \citet{park2025flow} propose Flow Q-learning (FQL), an offline RL approach that trains a flow-based policy and distills it into a one-step policy for fine-tuning, inspired by diffusion Q-learning.
Flow Policy Optimization (FPO)~\cite{mcallister2025flow} reframes policy optimization as minimizing an advantage-weighted conditional flow matching loss.
\citet{wagenmaker2025steering} perform reinforcement learning in the latent noise space rather than updating policy parameters.
\citet{zhang2025reinflow} propose a policy gradient–based method that injects learnable noise during inference and reformulates the inference process as a Markov decision process.

While these policy gradient–based approaches demonstrate strong empirical performance, they lack a unified framework that characterizes their common structure and provides theoretical guarantees for transferring improvements from training-time stochastic inference to deployment-time deterministic inference.
This work addresses this gap.

At the time of finalizing this paper, we became aware of a concurrent work~\cite{chen2025pi}, which proposes two approaches for flow matching policy optimization.
One approach, Flow-SDE, shares the FMPO setting of SDE inference with a PPO surrogate, while Flow-Noise uses learnable noise injection within the denoising process.

\clearpage

\section{Algorithms}
\label{sec:algorithms}

The algorithm summarizes rollout collection and actor updates using a supplied advantage estimator.

\begin{algorithm}[H]

\caption{Flow Matching Policy Optimization with ChainPPO (FMPO-CPPO)}
\label{algo:FMPO_chainppo}
\begin{algorithmic}[1]
\Require Environment MDP $\mathcal{M}_{\rm env}=(\mathcal{S},\mathcal{A},P,R,\gamma)$; pretrained velocity $u_\theta$; denoising steps $K$; flow times $\tau_k=k/K$, $k=0,\ldots,K$; noise schedule $\lambda_\tau$; chain-level clip $\epsilon_{\rm c}$; step-wise clip $\epsilon_{\rm s}$; auxiliary coefficient $\alpha$; advantage estimator $\widehat{\mathcal{A}}$; learning rate $\eta$; rollout horizon $T_{\rm env}$; rollouts per iteration $N$; training iterations $M$; PPO epochs $E$; action-chunk dimension $D=Hd_a$.
\For{training iteration $m=1,\ldots,M$}
    \State $\theta_{\rm old}\leftarrow \theta$, $\mathcal{D}\leftarrow \emptyset$
    \For{rollout $i=1,\ldots,N$}
        \State Reset environment and observe $s_0^i$
        \For{$t=0,\ldots,T_{\rm env}-1$}
            \State Sample initial noise $a_t^{0,i}\sim N(0,I)$
            \For{$k=1,\ldots,K$}
                \State Sample $a_t^{k,i}\sim \bar{\pi}_{\theta_{\rm old}}(\cdot\mid s_t^i,\tau_{k-1},a_t^{k-1,i})$ by the Euler--Maruyama kernel evaluated at flow time $\tau_{k-1}$ with noise level $\lambda_{\tau_{k-1}}$
                \State Store $(s_t^i,a_t^{k-1,i},a_t^{k,i},k)$ in $\mathcal{D}$
            \EndFor
            \State Execute $a_t^i\leftarrow a_t^{K,i}$ in the environment
            \State Observe $s_{t+1}^i\sim P(\cdot\mid s_t^i,a_t^i)$ and reward $R_t^i=R(s_t^i,a_t^i)$
            \State Store $(s_t^i,a_t^{0:K,i},R_t^i)$ in $\mathcal{D}$
        \EndFor
    \EndFor
    \State Estimate environment-step advantages $\{\widehat A_t^i\}\leftarrow \widehat{\mathcal{A}}(\mathcal{D},\{R_t^i\})$ and hold them fixed during actor updates
    \For{PPO epoch $e=1,\ldots,E$}
        \State Compute normalized step-wise ratios
        \[
        \tilde r_{t,k}^i(\theta)
        =
        \left(
        \frac{
        \bar{\pi}_{\theta}(a_t^{k,i}\mid s_t^i,\tau_{k-1},a_t^{k-1,i})
        }{
        \bar{\pi}_{\theta_{\rm old}}(a_t^{k,i}\mid s_t^i,\tau_{k-1},a_t^{k-1,i})
        }
        \right)^{1/D}.
        \]
        \State Compute normalized chain-level ratios
        \[
        \tilde\rho_t^i(\theta)
        =
        \prod_{k=1}^K \tilde r_{t,k}^i(\theta).
        \]
        \State Form the chain-level PPO surrogate
        \[
        \mathcal{J}_{\rm chain}(\theta)
        =
        \widehat{\mathbb{E}}_{i,t}
        \left[
        \min\left\{
        \tilde\rho_t^i(\theta)\widehat A_t^i,\,
        \operatorname{Clip}\left(\tilde\rho_t^i(\theta),1-\epsilon_{\rm c},1+\epsilon_{\rm c}\right)\widehat A_t^i
        \right\}
        \right].
        \]
        \State Form the auxiliary step-wise PPO surrogate
        \[
        \mathcal{J}_{\rm step}(\theta)
        =
        \widehat{\mathbb{E}}_{i,t,k}
        \left[
        \min\left\{
        \tilde r_{t,k}^i(\theta)\widehat A_t^i,\,
        \operatorname{Clip}\left(\tilde r_{t,k}^i(\theta),1-\epsilon_{\rm s},1+\epsilon_{\rm s}\right)\widehat A_t^i
        \right\}
        \right].
        \]
        \State Update $\theta \leftarrow \theta + \eta \nabla_\theta\bigl(\mathcal{J}_{\rm chain}(\theta)+\alpha\mathcal{J}_{\rm step}(\theta)\bigr)$
    \EndFor
\EndFor
\end{algorithmic}

\end{algorithm}

\clearpage
\section{\texorpdfstring{Proof of \cref{thm:correctness}}{Proof of Theorem \ref{thm:correctness}}}
\label{sec:proof_correctness}

We compare the learned ODE and SDE with the reference path generated by the true velocity.
\cref{sec:proof_correct_preliminaries} provides a Wasserstein stability bound, a density regularity bound, and a Wasserstein-to-TV interpolation inequality for the ODE comparison, together with a Girsanov bound for the SDE comparison.
\cref{sec:main_proof_correctness} combines these bounds to prove \cref{thm:correctness}.

\subsection{Preliminaries}
\label{sec:proof_correct_preliminaries}
We start with the definition of total variation distance.
\begin{definition}[Total Variation Distance]
\label{def:tv_distance}
Let $P$ and $Q$ be probability measures on $\mathbb{R}^d$.
Their total variation (TV) distance is
\begin{align*}
    \text{TV}(P,Q):=\sup_{A\in\mathcal{B}(\mathbb{R}^d)}|P(A)-Q(A)|.
\end{align*}
If they have densities $p$ and $q$ with respect to Lebesgue measure, then
\begin{align*}
    \text{TV}(P,Q)=\frac12\int_{\mathbb{R}^d}|p(x)-q(x)|\,\dd x.
\end{align*}
\end{definition}

The SDE comparison uses the following consequence of Girsanov's theorem \cite{karatzas2014brownian}.
We include a localization argument for globally Lipschitz drifts.
\begin{lemma}[Girsanov's Theorem, \cite{karatzas2014brownian}]
\label{lem:girsanov}
Let $Z$ and $Z'$ be the unique non-explosive solutions of
\begin{align*}
    \dd Z_t&=b(Z_t,t)\,\dd t+\sigma(t)\,\dd W_t,\qquad Z_0\sim p_0,\\
    \dd Z'_t&=b'(Z'_t,t)\,\dd t+\sigma(t)\,\dd W_t,\qquad Z'_0\sim p_0,
\end{align*}
where $\sigma:[0,T]\to(0,\infty)$ is continuous.
Let $p_t,p'_t$ denote their marginal laws and $\P$ the path law of $Z$.
Assume either Novikov's condition
\begin{align*}
    \E_{\P}\left[\exp\left(\frac12\int_0^T
    \frac{\|(b-b')(Z_t,t)\|_2^2}{\sigma^2(t)}\,\dd t\right)\right]<\infty
\end{align*}
holds, or both drifts are continuous, globally Lipschitz in space uniformly in time, and have at most linear growth uniformly in time.
Then
\begin{align*}
    \text{KL}(p_T||p'_T)
    \leq\frac12\int_0^T\E_{Z_t\sim p_t}
    \left[\frac{\|(b-b')(Z_t,t)\|_2^2}{\sigma^2(t)}\right]\,\dd t.
\end{align*}
\end{lemma}
\begin{proof}
Under Novikov's condition, Girsanov's theorem gives the relative entropy between the path laws.
The stated bound follows by applying the data processing inequality to the terminal evaluation map.

For globally Lipschitz drifts, let $\chi_n$ be a smooth function with $0\le\chi_n\le1$, equal to $1$ on the ball of radius $n$ and supported in the ball of radius $2n$.
Define $b'_n=b+\chi_n(b'-b)$, and let $p'_{n,T}$ be the terminal law under drift $b'_n$.
Since $b'_n-b$ is bounded and $\sigma$ has a positive lower bound, Novikov's condition holds for this comparison.
Thus,
\begin{align*}
    \text{KL}(p_T||p'_{n,T})
    &\leq\frac12\int_0^T\E_{p_t}
    \left[\frac{\chi_n^2(Z_t)\|(b-b')(Z_t,t)\|_2^2}{\sigma^2(t)}\right]\,\dd t\\
    &\leq\frac12\int_0^T\E_{p_t}
    \left[\frac{\|(b-b')(Z_t,t)\|_2^2}{\sigma^2(t)}\right]\,\dd t.
\end{align*}
Couple the processes with drifts $b'_n$ and $b'$ using the same initial variable and Brownian motion.
Their paths agree until they leave the ball of radius $n$, by pathwise uniqueness.
Since the process with drift $b'$ is non-explosive,
\begin{align*}
    \text{TV}(p'_{n,T},p'_T)
    \leq\Pr\left(\sup_{0\le t\le T}\|Z'_t\|_2\ge n\right)\longrightarrow0.
\end{align*}
Lower semicontinuity of relative entropy then gives the same bound for $p'_T$.
\end{proof}

For the ODE comparison, we first recall the Wasserstein stability estimate for flow matching.
For $r\in\{1,2\}$, write
\begin{align*}
    W_r(p,q):=\inf_{\pi\in\Pi(p,q)}
    \left(\int\|x-y\|_2^r\,\pi(\dd x,\dd y)\right)^{1/r},
\end{align*}
where $\Pi(p,q)$ is the set of couplings of $p$ and $q$.
The following is the time-rescaled stability estimate of \citet[Theorem~1]{benton2023error}.
We give the coupling proof for completeness.
\begin{lemma}[Wasserstein Bound For Flow Matching]
\label{lem:wasserstein_bound_ode}
Let $u,v\in C([0,T];C^1(\R^d;\R^d))$, and let $X_t,Y_t$ be their ODE flows with the same initial law of finite second moment.
Assume that both flows exist on $[0,T]$, that $v(\cdot,t)$ is globally $L$-Lipschitz, and that
\begin{align*}
    \int_0^T\E_{X_t\sim p_t^\star}
    [\|u(X_t,t)-v(X_t,t)\|_2^2]\,\dd t\le\epsilon^2,
\end{align*}
where $p_t^\star$ is the marginal law of $X_t$.
If $p_T^v$ is the law of $Y_T$, then
\begin{align*}
    W_1(p_T^\star,p_T^v)\le W_2(p_T^\star,p_T^v)
    \le\sqrt T e^{LT}\epsilon.
\end{align*}
\end{lemma}
\begin{proof}
Couple the two ODE flows using $X_0=Y_0$.
By the Lipschitz condition on $v$ and Gronwall's inequality,
\begin{align*}
    \|Y_T-X_T\|_2
    \le\int_0^T e^{L(T-s)}\|v(X_s,s)-u(X_s,s)\|_2\,\dd s.
\end{align*}
Cauchy--Schwarz gives
\begin{align*}
    \E\|Y_T-X_T\|_2^2
    &\le Te^{2LT}\int_0^T\E_{p_s^\star}
    [\|v(X_s,s)-u(X_s,s)\|_2^2]\,\dd s\\
    &\le Te^{2LT}\epsilon^2.
\end{align*}
The conclusion follows from the definition of $W_2$ and $W_1\le W_2$.
\end{proof}

To convert this Wasserstein bound into a TV bound, we control the spatial variation of the terminal densities.
For a density $p$, define
\begin{align*}
    S(p):=\|\nabla p\|_{L^1}
    =\int_{\R^d}\|\nabla p(x)\|_2\,\dd x.
\end{align*}
\begin{lemma}[Density Regularity Under An ODE Flow]
\label{lem:density_regularity_ode}
Let $v\in C([0,T];C^2(\R^d;\R^d))$ satisfy
$\|D_xv\|_{\mathrm{op}}\le L$ and $\|\nabla_x\operatorname{div}v\|_2\le H$.
Let $p_t^v$ be the density induced by its ODE flow from $p_0=\mathcal{N}(0,I_d)$.
Then
\begin{align*}
    S(p_T^v)\le e^{LT}\E\|G\|_2
    +H\int_0^T e^{L(T-s)}\,\dd s,
    \qquad G\sim\mathcal{N}(0,I_d).
\end{align*}
\end{lemma}
\begin{proof}
The Lipschitz condition and time continuity give a global flow $F_t$ that is $C^2$ in the initial state.
The change-of-variables formula gives a positive $C^1$ density $p_t^v$.
Let $g_t=\nabla\log p_t^v$.
Differentiating the change-of-variables formula along the flow yields
\begin{align*}
    \frac{\dd}{\dd t}g_t(F_t(a))
    =-D_xv(F_t(a),t)^\top g_t(F_t(a))
    -\nabla_x\operatorname{div}v(F_t(a),t).
\end{align*}
Since $g_0(a)=-a$, Gronwall's inequality gives
\begin{align*}
    \|g_T(F_T(a))\|_2
    \le e^{LT}\|a\|_2+H\int_0^T e^{L(T-s)}\,\dd s.
\end{align*}
Integrating over $a\sim p_0$ and using
$S(p_T^v)=\E_{p_T^v}\|g_T\|_2$ proves the claim.
\end{proof}

The next lemma is the standard smoothing argument that relates Wasserstein distance to TV for densities with integrable gradients.
\begin{lemma}[Wasserstein-To-TV Interpolation]
\label{lem:wasserstein_tv}
Let $p,q\in W^{1,1}(\R^d)$ be probability densities with finite first moments.
Then
\begin{align*}
    \text{TV}(p,q)
    \le\sqrt{\frac2\pi}\sqrt{W_1(p,q)\bigl(S(p)+S(q)\bigr)}.
\end{align*}
\end{lemma}
\begin{proof}
Let $\gamma_h$ be the density of $\mathcal{N}(0,h^2I_d)$, let $G\sim\mathcal{N}(0,I_d)$, and set $c=\sqrt{2/\pi}$.
We first prove the estimates for smooth densities; convolution approximation extends them to $W^{1,1}$, since it preserves the gradient bounds and converges in both $L^1$ and $W_1$.
The fundamental theorem of calculus, translation invariance, and
$\E|G^\top z|=c\|z\|_2$ give
\begin{align*}
    \|p-p*\gamma_h\|_{L^1}
    \le h\int_{\R^d}\E|G^\top\nabla p(x)|\,\dd x
    =chS(p).
\end{align*}
The same bound holds for $q$.
Moreover, for any coupling $\pi\in\Pi(p,q)$,
\begin{align*}
    \|p*\gamma_h-q*\gamma_h\|_{L^1}
    &\le\int\|\gamma_h(\cdot-x)-\gamma_h(\cdot-y)\|_{L^1}\,\pi(\dd x,\dd y)\\
    &\le\frac c h\int\|x-y\|_2\,\pi(\dd x,\dd y).
\end{align*}
Here the second inequality follows from
$\|\partial_e\gamma_h\|_{L^1}=c/h$ for every unit vector $e$.
Taking the infimum over couplings and applying the triangle inequality yields
\begin{align*}
    \text{TV}(p,q)
    \le\frac c2\left(h\bigl(S(p)+S(q)\bigr)+\frac{W_1(p,q)}h\right).
\end{align*}
Optimizing over $h>0$ proves the stated bound.
\end{proof}

\subsection{Main Proof of \texorpdfstring{\cref{thm:correctness}}{Theorem \ref{thm:correctness}}}
\label{sec:main_proof_correctness}

\begin{theorem}
[Formal Version Of \cref{thm:correctness}]
\label{thm:correctness_re}
Let $0<T<1$ and assume that $p_1$ has a finite second moment.
Let $u\in C([0,T];C^1(\R^d;\R^d))$ be the true velocity of the independent linear path, and assume that $u(\cdot,t)$ is globally Lipschitz uniformly in $t$.
Let $u_\theta\in C([0,T];C^2(\R^d;\R^d))$ be the estimated velocity satisfying
\begin{align*}
    \|D_xu_\theta(x,t)\|_{\mathrm{op}}\le L,\qquad
    \|\nabla_x\operatorname{div}u_\theta(x,t)\|_2\le H
\end{align*}
for all $x\in\R^d$ and $t\in[0,T]$, where $L,H<\infty$ are uniform over the models under consideration.
Let $W_t$ be a standard Wiener process and let the noise schedule $\lambda:[0,T]\to(0,\infty)$ be continuous.
For parameter $\theta$, we define three processes $X_t$, $Y_t^\theta$, and $Z_t^\theta$ by
\begin{align*}
    \dd X_t 
    = & ~ u(X_t,t) \dd t,
    \quad
    \dd Y_t^\theta
    =  u_\theta(Y_t^\theta,t) \dd t, \\
    \dd Z_t^\theta
    = & ~ \begin{aligned}[t]
        &\left[\left(1+\frac{\lambda_t t}{2(1-t)}\right)u_\theta(Z_t^\theta,t)\right.\\
        &\left.\quad-\frac{\lambda_t}{2(1-t)}Z_t^\theta\right]\dd t
        +\sqrt{\lambda_t}\dd W_t.
    \end{aligned}
\end{align*}

We assume that $u_\theta$ is close to $u$,
\begin{align}
\label{eq:similar_velocity}
    \int_0^T \E_{
    X_t \sim p^\star_t}
    [\|u(X_t,t)-u_\theta(X_t,t)\|_2^2]
    \dd t
    \leq \epsilon^2.
\end{align}
Let $p_t^\star$, $p_t$, and $q_t$ denote the probability paths induced by $X_t$, $Y_t^\theta$, and $Z_t^\theta$, respectively.
All processes share the same initial distribution $p_0=\mathcal{N}(0,I_d)$.
Then the terminal distributions satisfy
\begin{align*}
    \text{TV}(p_T||q_T)
    \leq C_{\mathrm{ODE}}\sqrt{\epsilon}
    +\frac{\sqrt{C_{\lambda,T}}}{2}\epsilon
    =O(\sqrt{\epsilon})
\end{align*}
as $\epsilon\to0$, where $C_{\mathrm{ODE}}$ and $C_{\lambda,T}$ depend only on $d,T,L,H$ and the noise schedule, and are defined in \cref{sec:main_proof_correctness}.
\end{theorem}

\begin{proof}
We first compare the learned ODE with the reference ODE.
Let $\rho_T$ be the law of $TX_1$, let $\gamma_h$ be the density of $\mathcal{N}(0,h^2I_d)$, and write $c_d=\E\|G\|_2$ for $G\sim\mathcal{N}(0,I_d)$.
The independent linear path gives $p_T^\star=\rho_T*\gamma_{1-T}$.
Therefore,
\begin{align*}
    S(p_T^\star)
    \le\|\nabla\gamma_{1-T}\|_{L^1}
    =\frac{c_d}{1-T}.
\end{align*}
By \cref{lem:density_regularity_ode}, the learned ODE density satisfies
\begin{align*}
    S(p_T)\le e^{LT}c_d+H\int_0^T e^{L(T-s)}\,\dd s.
\end{align*}
Define
\begin{align*}
    B_T&:=\frac{c_d}{1-T}+e^{LT}c_d
    +H\int_0^T e^{L(T-s)}\,\dd s,\\
    C_{\mathrm{ODE}}&:=\sqrt{\frac2\pi}\,T^{1/4}e^{LT/2}\sqrt{B_T}.
\end{align*}
All terminal laws have finite second moments by the linear-growth bounds on the velocities.
Combining \cref{lem:wasserstein_bound_ode,lem:wasserstein_tv} gives
\begin{align*}
    \text{TV}(p_T^\star,p_T)
    &\le\sqrt{\frac2\pi}\sqrt{W_1(p_T^\star,p_T)B_T}\\
    &\le C_{\mathrm{ODE}}\sqrt{\epsilon}.
\end{align*}

We next compare the learned SDE with the reference path.
For the true velocity $u$, \eqref{eq:velocity_score} gives
\begin{align*}
    \nabla\log p_t^\star(x)=\frac{t\,u(x,t)-x}{1-t},
\end{align*}
with the value at $t=0$ given by continuity.
Thus, the reference ODE has a dual stochastic process
\begin{align*}
    \dd\widetilde X_t
    ={}&\left[\left(1+\frac{\lambda_t t}{2(1-t)}\right)u(\widetilde X_t,t)
    -\frac{\lambda_t}{2(1-t)}\widetilde X_t\right]\,\dd t
    +\sqrt{\lambda_t}\,\dd W_t,
\end{align*}
with initial law $p_0$ and the same marginal laws $p_t^\star$.
Indeed, its Fokker--Planck equation reduces to the reference continuity equation by \eqref{eq:velocity_score}.
Uniqueness of the Fokker--Planck equation for globally Lipschitz drifts identifies its marginal laws with $p_t^\star$.
Both SDE drifts are globally Lipschitz with linear growth, since $T<1$ and $\lambda$ is continuous and positive on $[0,T]$.
Consequently, \cref{lem:girsanov} applies and gives
\begin{align*}
    \text{KL}(p_T^\star||q_T)
    \le{}&\frac12\int_0^T\frac1{\lambda_t}
    \left(1+\frac{\lambda_t t}{2(1-t)}\right)^2
    \E_{p_t^\star}\|(u-u_\theta)(X_t,t)\|_2^2\,\dd t.
\end{align*}
Define
\begin{align*}
    C_{\lambda,T}
    :=\operatorname*{ess\,sup}_{t\in[0,T]}
    \frac1{\lambda_t}\left(1+\frac{\lambda_t t}{2(1-t)}\right)^2.
\end{align*}
This coefficient is finite under the noise-schedule assumption.
For constant noise $\lambda_t\equiv\lambda>0$, it equals
$\lambda^{-1}(1+\lambda T/(2(1-T)))^2$.
Using \eqref{eq:similar_velocity} and Pinsker's inequality, we obtain
\begin{align*}
    \text{KL}(p_T^\star||q_T)&\le\frac{C_{\lambda,T}}2\epsilon^2,\\
    \text{TV}(p_T^\star,q_T)&\le\frac{\sqrt{C_{\lambda,T}}}{2}\epsilon.
\end{align*}
Finally, the triangle inequality yields
\begin{align*}
    \text{TV}(p_T||q_T)
    &\le\text{TV}(p_T,p_T^\star)+\text{TV}(p_T^\star,q_T)\\
    &\le C_{\mathrm{ODE}}\sqrt{\epsilon}
    +\frac{\sqrt{C_{\lambda,T}}}{2}\epsilon
    =O(\sqrt{\epsilon}).
\end{align*}
The constants are independent of $\theta$ and $\epsilon$ when $L,H$ and the noise schedule are fixed.
This completes the proof.
\end{proof}

\begin{remark}[Verification Of The Regularity Conditions]
\label{rem:correctness_regularity}
The reference-velocity assumption holds for Gaussian targets and for bounded targets on every fixed interval $[0,T]$ with $T<1$.
For Gaussian targets, the reference velocity is affine in $x$, as in \cref{sec:validate_corr}.
For bounded targets with $\|X_1\|_2\le R_1$, define $m_t(x)=\E[X_1\mid X_t=x]$ along the independent linear path.
Gaussian conditioning gives
\begin{align*}
    u(x,t)&=\frac{m_t(x)-x}{1-t},\\
    D_xm_t(x)&=\frac{t}{(1-t)^2}\operatorname{Cov}(X_1\mid X_t=x).
\end{align*}
Since the conditional covariance has operator norm at most $R_1^2$, the reference velocity is smooth in space and satisfies
\begin{align*}
    \|D_xu(x,t)\|_{\mathrm{op}}
    \le\frac1{1-T}+\frac{TR_1^2}{(1-T)^3},\qquad t\le T.
\end{align*}

The learned-velocity assumptions hold for each fixed finite-parameter SiLU MLP used in the synthetic experiments.
To see this, write $h_\ell=\sigma(W_\ell h_{\ell-1}+b_\ell)$ and let
$\beta_1=\sup_z|\sigma'(z)|<\infty$ and $\beta_2=\sup_z|\sigma''(z)|<\infty$.
If $J_\ell,K_\ell$ bound the first and second spatial derivatives of $h_\ell$, respectively, the chain rule gives
\begin{align*}
    J_\ell&\le\beta_1\|W_\ell\|_{\mathrm{op}}J_{\ell-1},\\
    K_\ell&\le\beta_2\|W_\ell\|_{\mathrm{op}}^2J_{\ell-1}^2
    +\beta_1\|W_\ell\|_{\mathrm{op}}K_{\ell-1}.
\end{align*}
Starting from $J_0=1,K_0=0$, finite depth and finite weights give finite derivative bounds, and the linear output layer preserves them.
A bound $K$ on the second derivative of the output gives $\|\nabla\operatorname{div}u_\theta\|_2\le dK$.
For a fixed architecture with uniformly bounded layer weights, these recursions give uniform $L,H$ over the model family.
For a single model, the bound remains valid with model-dependent $L_\theta,H_\theta$ and $C_{\mathrm{ODE},\theta}$.
\end{remark}

\section{\texorpdfstring{Proof of \cref{thm:improve_trans}}{Proof of Theorem \ref{thm:improve_trans}}}
\label{sec:proof_improve_trans}

This section provides the proof of \cref{thm:improve_trans}.

\begin{theorem}[\cref{thm:improve_trans} Restated: Transfer of improvement from SDE to ODE]
\label{thm:improve_trans——re}
Let $r: \R^d \rightarrow \R$ be the terminal reward.
Assume the terminal reward is bounded, i.e., $\abs{r(x)} < R < \infty$ for any $x \in \R^d$.
Fix early stopping time $0<T<1$ and noise level $\lambda_t >0$.
For each parameter $\theta$, we define two processes:
\begin{itemize}
    \item The SDE (training-time) inference process $Z^\theta=(Z^\theta_{t})_{t
    \in [0,T]}$ by 
    \begin{align*}
        \dd Z^\theta_t
        = & ~ [(1+\frac{\lambda_t}{2} \frac{t}{1-t}) u_\theta(Z^\theta_t, t)
        - (\frac{\lambda_t}{2} \frac{1}{1-t}) Z^\theta_t] \dd t
        + \sqrt{\lambda_t} \dd W_t, \quad 
        Z^\theta_0 \sim p_0.
    \end{align*}
    \item The ODE (deployment-time) inference process $Y^\theta=(Y^\theta_{t})_{t
    \in [0,T]}$ by 
    \begin{align*}
        \dd Y_t
        = & ~  u_\theta
        (Y^\theta_t, t)
        \dd t, \quad 
        Y^\theta_0 \sim p_0.
    \end{align*}
\end{itemize}
Define the corresponding performances
\begin{align*}
    J_{\text{SDE}}(\theta)
    := \E[r(Z^\theta_T)],
    \quad
    J_{\text{ODE}}(\theta)
    := \E[r(Y^\theta_T)].
\end{align*}
Let $\{\theta_k\}_{k\geq0}$ be a sequence of parameters produced by a reinforcement learning algorithm that satisfies the following  one-step improvement condition on the SDE performance: for some $\Delta_k>0$,
\begin{align*}
    J_{\text{SDE}}(\theta_{k+1})
    \geq J_{\text{SDE}}(\theta_{k}) + \Delta_k.
\end{align*}
Assume that the hypotheses of \cref{thm:correctness_re} hold for every $\theta_k$ with the same constants $L,H$ and noise schedule.
Then the ODE performance satisfies the transfer improvement bound
\begin{align*}
    J_{\text{ODE}}(\theta_{k+1})
    \geq J_{\text{ODE}}(\theta_{k}) + \Delta_k - 2R(\delta_k + \delta_{k+1}),
\end{align*}
where $\delta_k := C_{\mathrm{ODE}}\sqrt{\epsilon_k}+\frac{\sqrt{C_{\lambda,T}}}{2}\epsilon_k=O(\sqrt{\epsilon_k})$ is induced by the flow-matching error $\epsilon_k := \sqrt{\mathcal{L}_{\mathrm{FM},T}(\theta_k)}$.
\end{theorem}

\begin{proof}
    Let $q_t^k$, and $p_t^k$ denote the probability paths induced by $Z_t^{\theta_k}$, and $Y_t^{\theta_k}$, respectively. 
    Then from the  correctness for flow RL via diffusion-flow duality (\cref{thm:correctness_re}), we have for any $k\geq0$
    \begin{align}
    \label{eq:correct_pro1}
        \text{TV}(p_T^k||q_T^k)
        \leq \delta_k.
    \end{align}
    Here $\delta_k=C_{\mathrm{ODE}}\sqrt{\epsilon_k}+\frac{\sqrt{C_{\lambda,T}}}{2}\epsilon_k=O(\sqrt{\epsilon_k})$ is induced by the flow-matching error
    \begin{align*}
        \epsilon_k^2 := \int_0^T \E_{
        X_t \sim p^\star_t}
        [\|u(X_t,t)-u_{\theta_k}(X_t,t)\|_2^2]
        \dd t,
    \end{align*}
    where $u$ is the true (constructed) velocity and $p^*$ is the probability path induced by $u$.
    
    A standard property of total variation distance is that for any two distributions
    $P,Q$ on $\R^d$ and any bounded measurable function $f: \R^d \rightarrow \R$,
    \begin{align*}
        \abs{\E_{X \sim P}[f(X)] - \E_{X \sim Q}[f(X)]}
        \leq 2 \|f\|_\infty
        \text{TV}(P||Q).
    \end{align*}
    Let $f$ be the terminal reward function $r$ (bounded by $R$). 
    Then by \eqref{eq:correct_pro1}, we have
    \begin{align}
        \abs{\E_{X \sim p_T^k}[r(X)] - \E_{X \sim q_T^k}[r(X)]}
        \leq & ~ 2 R \cdot
        \text{TV}(p_T^{k}||
        q_T^{k})
        \leq 2R\delta_k, 
        \label{eq:k_performance_gap}\\
        \abs{\E_{X \sim p_T^{k+1}}[r(X)] - \E_{X \sim q_T^{k+1}}[r(X)]}
        \leq & ~ 2 R \cdot
        \text{TV}(p_T^{k+1}||
        q_T^{k+1})
        \leq 2R\delta_{k+1}
        \label{eq:k_1_performance_gap}.
    \end{align}
    Next, by the one-step improvement of the SDE performance assumption,
    \begin{align}
    \label{eq:one_step_im}
        J_{\text{SDE}}(\theta_{k+1})
        \geq J_{\text{SDE}}(\theta_{k}) + \Delta_k
        \quad \text{for some} ~ \Delta_k \geq 0.
    \end{align}
    We now prove the one-step improvement bound for ODE performance by applying these inequalities:
    \begin{align*}
        J_{\text{ODE}}(\theta_{k+1}) 
        \geq & ~ 
        J_{\text{SDE}}(\theta_{k+1})
        - 2R \delta_{k+1},
        \annot{By \eqref{eq:k_1_performance_gap}} \\
        \geq & ~ J_{\text{SDE}}(\theta_{k})
        + \Delta_k
        - 2R \delta_{k+1},
        \annot{By \eqref{eq:one_step_im}} \\
        \geq & ~ J_{\text{ODE}}(\theta_{k})
        + \Delta_k
        - 2R (\delta_{k+1} + \delta_k).
        \annot{By \eqref{eq:k_performance_gap}}
    \end{align*}
    This completes the proof.
\end{proof}

\section{Connection between ChainPPO and the Composite-MDP Objective}
\label{sec:unbias_cppo}

We first show that treating the full inference chain as a macro-action does not change the original objective.
We then analyze the policy-gradient directions induced by the chain-level and step-wise PPO surrogates before clipping.
Throughout this section, $J_{\rm chain}$ and $J_{\rm step}$ refer to the corresponding unclipped surrogate objectives.
The clipped surrogates define the actor updates in \cref{algo:FMPO_chainppo}.

\begin{proposition}[Macro-action objective equivalence]
\label{prop:macro_action_equiv}
Consider the composite MDP in \cref{sec:fm_policy_op}.
At each environment step $t$, let
\[
    c_t := (a_t^0,a_t^1,\ldots,a_t^K)
\]
denote the full stochastic inference chain, where $a_t^0 \sim \mu_0=N(0,I)$ and
\[
    \Pi_\theta(c_t \mid s_t)
    =
    \mu_0(a_t^0)
    \prod_{k=1}^K
    p_{\tau_{k-1}}^\theta(a_t^k \mid a_t^{k-1},s_t).
\]
Define the macro-action reward and transition kernel by
\[
    R_{\rm chain}(s_t,c_t) := R(s_t,a_t^K),
    \qquad
    P_{\rm chain}(s_{t+1}\mid s_t,c_t)
    :=
    P(s_{t+1}\mid s_t,a_t^K).
\]
Let
\[
    \mathcal J_{\rm macro}(\theta)
    :=
    \E_{\Pi_\theta,P_{\rm chain}}
    \left[
        \sum_{t\ge 0}\gamma^t R_{\rm chain}(s_t,c_t)
    \right]
\]
be the expected return of this macro-action MDP. Then
\[
    \mathcal J_{\rm macro}(\theta)
    =
    \mathcal J_{\rm com}(\theta)
    =
    \mathcal J_{\rm env}(\theta).
\]
\end{proposition}

\begin{proof}
By definition, the macro-action $c_t$ generates the executed environment action $a_t=a_t^K$.
Thus the macro-action reward satisfies
\[
    R_{\rm chain}(s_t,c_t)
    =
    R(s_t,a_t^K),
\]
which is exactly the reward assigned to the terminal inner step of the composite MDP.
The macro-action transition kernel also matches the environment transition after executing $a_t^K$:
\[
    P_{\rm chain}(s_{t+1}\mid s_t,c_t)
    =
    P(s_{t+1}\mid s_t,a_t^K).
\]
Therefore, marginalizing over the intermediate inference variables
$a_t^0,\ldots,a_t^{K-1}$ gives the same distribution over environment trajectories
$(s_t,a_t^K,s_{t+1})_{t\ge 0}$ as the original flow policy.
Hence the macro-action MDP and the environment MDP have the same expected return:
\[
    \mathcal J_{\rm macro}(\theta)
    =
    \mathcal J_{\rm env}(\theta).
\]
The equality
$\mathcal J_{\rm com}(\theta)=\mathcal J_{\rm env}(\theta)$
follows from the construction of the composite MDP in \cref{sec:fm_policy_op}.
Thus
\[
    \mathcal J_{\rm macro}(\theta)
    =
    \mathcal J_{\rm com}(\theta)
    =
    \mathcal J_{\rm env}(\theta).
\]
\end{proof}

\section{Additional Details of Experiments}
\label{sec:implementation}
\subsection{Validating Correctness (\texorpdfstring{\cref{thm:correctness}}{Theorem~\ref{thm:correctness}})}
\textbf{Model and Training Details.} We approximate the velocity field $u_\theta(x,t)\colon\mathbb{R}^d\times[0,T]\to\mathbb{R}^d$ with a multilayer perceptron. The input $(x,t)\in\mathbb{R}^{d+1}$ is formed by concatenating the spatial state $x\in\mathbb{R}^d$ with the scalar time $t\in[0,T]$. The network has three hidden layers of width $96$ with SiLU activations $\sigma(z)=z\,\mathrm{sigmoid}(z)$, i.e. the architecture is
$$
(d{+}1)\xrightarrow{\text{Linear+SiLU}} 96 \xrightarrow{\text{Linear+SiLU}} 96 \xrightarrow{\text{Linear+SiLU}} 96 \xrightarrow{\text{Linear}} d,
$$
so that the output layer is linear and matches the dimension of the velocity field. 

We train $u_\theta$ by marginal flow-matching regression, using the closed-form target below, along the linear interpolation path $X_t=(1-t)X_0+tX_1$ with independent coupling $X_0\sim\mathcal{N}(0,I_d)$ and $X_1\sim\mathcal{N}(\mu,\Sigma)$. At each iteration we draw a fresh mini-batch of size $512$ by sampling $t\sim\mathcal{U}(0,T)$ with $T=0.99$, independent $X_0,X_1$ from the marginals above, and forming $X_t$ on-the-fly; the empirical loss is the per-coordinate mean squared error between $u_\theta(X_t,t)$ and the target velocity. Because $(X_0,X_1)$ is jointly Gaussian, the Bayes-optimal velocity
$$
u^\star(x,t)=\mathbb{E}[X_1-X_0\mid X_t=x]=\mu+\bigl(t\Sigma+(t-1)I\bigr)\bigl((1-t)^2 I+t^2\Sigma\bigr)^{-1}(x-t\mu)
$$
is available in closed form; we use it as the regression target, which has the same minimizer and the same expected gradient as the standard CFM loss but lower variance, allowing us to control the training error $\epsilon$ precisely. Optimization uses Adam with learning rate $10^{-3}$, default momentum parameters $(\beta_1,\beta_2)=(0.9,0.999)$, $\epsilon_{\text{Adam}}=10^{-8}$, and no learning-rate schedule. The number of optimizer steps is varied over $20$ values evenly spaced between $100$ and $4000$ to sweep out a range of approximation errors $\epsilon$, and each configuration is repeated over $10$ independent seeds (controlling network initialization, data sampling, and integrator noise) to produce the error bars in \cref{fig:three}(a).

\subsection{Validating Performance Transfer (\texorpdfstring{\cref{thm:improve_trans}}{Theorem~\ref{thm:improve_trans}})}
\textbf{Model and Training Details.} The base policy $u_\theta(x,t)\colon\mathbb{R}^d\times[0,T]\to\mathbb{R}^d$ is parameterized by a multilayer perceptron that takes the concatenated input $(x,t)\in\mathbb{R}^{d+1}$ and returns the velocity field. The network has three hidden layers of width $128$ with SiLU activations $\sigma(z)=z\,\mathrm{sigmoid}(z)$ and a linear readout, i.e.
$$
(d{+}1)\xrightarrow{\text{Linear+SiLU}} 128 \xrightarrow{\text{Linear+SiLU}} 128 \xrightarrow{\text{Linear+SiLU}} 128 \xrightarrow{\text{Linear}} d,
$$ Both pretraining and fine-tuning phases use the time horizon $T=0.99$.
We pretrain $u_\theta$ by marginal flow-matching regression along the linear interpolation path $X_t=(1-t)X_0+tX_1$ with independent coupling $X_0\sim\mathcal{N}(0,I_d)$ and $X_1\sim\mathcal{N}(\mu_{\mathrm{tgt}},\Sigma_{\mathrm{tgt}})$, where $\mu_{\mathrm{tgt}}=(2.5,2.5)^\top$ and $\Sigma_{\mathrm{tgt}}=\Sigma_{\mathrm{base}}/\tau$ with $\Sigma_{\mathrm{base}}=\bigl(\begin{smallmatrix}0.25&0.18\\0.18&0.80\end{smallmatrix}\bigr)$ and reward temperature $\tau=3.0$. At each iteration we draw a fresh mini-batch of $2048$ pairs by sampling $t\sim\mathcal{U}(0,T)$ and independent $X_0,X_1$ from the marginals above, form $X_t$ on-the-fly, and minimize
$$
\mathcal{L}_{\mathrm{FM}}(\theta)=\mathbb{E}_{t,\,X_t}\bigl\Vert u_\theta(X_t,t)-u^\star(X_t,t)\bigr\Vert_2^{2},
$$ Pretraining uses Adam with learning rate $2\times 10^{-3}$, batch size $2048$, default momentum $(\beta_1,\beta_2)=(0.9,0.999)$, no schedule, and is run for $900$ steps; checkpoints are logged every $45$ steps to expose a family of pretrained velocities spanning a range of $\epsilon$ values, from which we select three starts (worst / middle / best) for the trajectory figure.

The reward is a temperature-scaled, log-clamped Mahalanobis kernel,
$$
r(x)=\exp\!\Bigl(\max\bigl\{-0.5\,\tau\,(x-\mu_{\mathrm{tgt}})^{\!\top}\Sigma_{\mathrm{base}}^{-1}(x-\mu_{\mathrm{tgt}}),\,L_{\min}\bigr\}\Bigr)\in(0,1],
$$
with $\tau=3.0$ and lower log-clamp $L_{\min}=-40$. With this choice, the implied target distribution is exactly $\mathcal{N}(\mu_{\mathrm{tgt}},\Sigma_{\mathrm{base}}/\tau)$, aligning the pretraining and fine-tuning objectives.

During RL fine-tuning, $\lambda=1$, the number of Euler--Maruyama steps is 100. Optimization uses Adam with learning rate $5\times 10^{-5}$, batch size $256$, gradient-norm clipping at $1.0$, and is run for $300$ steps; checkpoints are taken every $5$ steps to give the fine-grained reward trajectories shown in Fig.~\ref{fig:three}(c).
At each checkpoint we evaluate two test-time samplers from the same $X_0\sim\mathcal{N}(0,I_d)$ (common random numbers across configurations to stabilize differences): (i) the deterministic ODE $\mathrm{d}Y_t=u_\theta(Y_t,t)\,\mathrm{d}t$, and (ii) the same training-time SDE with $\lambda=1$. Both are integrated with $400$ Euler / Euler--Maruyama steps on $20000$ initial samples, yielding the reported $J_{\mathrm{ODE}}$ and $J_{\mathrm{SDE}}$. The flow-matching error $\epsilon=\sqrt{\int_0^T\!\mathbb{E}\,\Vert u_\theta-u^\star\Vert^2\,\mathrm{d}t}$ is estimated on $20000$ fresh $(t,X_t)$ samples and is used only as a diagnostic.

\subsection{FMPO-CPPO Evaluation}
We use the same settings as ReinFlow. Below are the implementation details of ReinFlow.

\textbf{ReinFlow Model and Training Details.} The ReinFlow policy is a conditional rectified-flow velocity field $u_\theta(x_t,t,o)$, where $x_t\in\mathbb{R}^{H d_a}$ is an action chunk, $t\in[0,1)$ is the flow time, and $o$ is the normalized observation history. It uses action chunking horizon $H=4$ and one observation frame. The flow network embeds $t$ with a sinusoidal embedding of dimension $16$, followed by a two-layer time MLP. For the Gym locomotion tasks, the policy trunk is a residual MLP with three hidden layers of width $512$ and ReLU activations, with a linear output of dimension $H d_a$:
$$
[x_t,\phi(t),o]\rightarrow 512\rightarrow 512\rightarrow 512\rightarrow H d_a .
$$
The critic is a separate residual MLP over observations with three hidden layers of width $256$ and Mish activations.

The policy is first pretrained by rectified-flow matching on offline action chunks. Given a dataset action chunk $x_1$ and Gaussian noise $x_0\sim\mathcal{N}(0,I)$, it samples $t\sim\mathcal{U}(0,1)$ and forms
$$
x_t=(1-t)x_0+t x_1,\qquad v=x_1-x_0 .
$$
The pretraining objective is the mean squared error
$$
\mathcal{L}_{\mathrm{RF}}(\theta)=\mathbb{E}\left[\|u_\theta(x_t,t,o)-(x_1-x_0)\|_2^2\right].
$$
Pretraining uses AdamW with learning rate $10^{-3}$, weight decay $10^{-6}$, batch size $128$, cosine warmup scheduling with minimum learning rate $10^{-4}$, and EMA decay $0.995$.

For online fine-tuning, ReinFlow uses PPO on the action chunks generated by Euler integration of the learned velocity field with $K=4$ denoising steps. It runs $40$ parallel environments, collects $500$ rollout steps per iteration, uses discount $\gamma=0.99$ and GAE parameter $\lambda=0.95$, and updates for $5$ PPO epochs. The PPO clipping coefficient is $\epsilon=0.01$, the value loss coefficient is $0.5$. Actor and critic are optimized with AdamW; the default Gym actor learning rate is $4.5\times 10^{-5}$ for Hopper, Ant, and Humanoid, and $4\times 10^{-4}$ for Walker2d, while the critic learning rate is $6.5\times10^{-4}$ except Walker2d, which uses $4\times10^{-3}$. Both use cosine warmup schedules and critic weight decay $10^{-5}$.

\textbf{Flow-SDE Training Details:} For Flow-SDE, all settings are the same as ReinFlow except the constant coefficient $a$ of noise level $\lambda_t$. Following ReinFlow, the PPO ratio is computed from the chain log-likelihood divided by $(K+1)Hd_a$. For each task, we search $a$ from $\{0.1,0.2,\ldots,0.9\}$ and select the best $a$ for each task. As a result, $a$ is 0.5,0.9,0.4,0.5 for Hopper, Walker2d, Ant, and Humanoid respectively.

\textbf{FMPO-CPPO Training Details:} For FMPO-CPPO, all settings are the same as ReinFlow unless otherwise specified. We additionally set the auxiliary weight $\alpha$ to 0.25 and the chained PPO clip coefficient to twice the PPO clip coefficient used by ReinFlow. The likelihood ratios are normalized by the action-chunk dimension $D=Hd_a$ as in \cref{sec:unified_framework}, so $\epsilon_{\rm c}=0.02$ and $\epsilon_{\rm s}=0.01$ apply to $\tilde\rho_t^{\rm chain}$ and $\tilde r_{t,k}$; the entropy bonus is normalized by the same $D$. We also search $a$ following the same procedure as Flow-SDE. The resulting $a$ is 0.8,0.6,0.3,0.8 for Hopper, Walker2d, Ant, and Humanoid respectively.

\textbf{Latency:} The table below is the training time per iteration for each algorithm. The unit is seconds.
\begin{table}[ht]
\centering
\caption{Training time per iteration in seconds.}
\begin{tabular}{lccc}
\toprule
\textbf{Environment} & \textbf{Flow-SDE} & \textbf{ReinFlow} & \textbf{FMPO-CPPO} \\
\midrule
Hopper   & $9.58 \pm 0.35$  & $12.28 \pm 1.81$ & $9.74 \pm 0.41$ \\
Walker2d & $14.95 \pm 1.50$ & $13.99 \pm 2.49$ & $15.12 \pm 1.47$ \\
Ant      & $16.40 \pm 1.96$ & $14.19 \pm 0.73$ & $16.46 \pm 2.01$ \\
Humanoid & $17.24 \pm 1.33$ & $19.65 \pm 0.59$ & $17.42 \pm 1.29$ \\
\bottomrule
\end{tabular}
\label{tab:performance_k4}
\end{table}
Overall, the training times of the three algorithms are similar.
We use a single NVIDIA A100 GPU with 40 GB of memory and a 10-core CPU for all training runs.

\subsection{Validating Correctness and Performance Transfer in the Image Generation Setting}
\textbf{Data and Reward.}
We use CIFAR-10, which contains $50000$ training and $10000$ test RGB images of size $32{\times}32$ in $10$ classes.
We reserve $10\%$ of the official training split for validation and rescale pixel values to $[-1,1]$ by $x\mapsto 2x-1$.
The test set is not used for pretraining or RL fine-tuning.
For a generated image $x$ conditioned on class $y\in\{0,\dots,9\}$, the reward is $r(x\mid y)=\phi(x)_y$, where $\phi(x)_y$ is the corresponding softmax probability from a frozen ResNet-18 pretrained on CIFAR-10.

\textbf{Velocity Model.}
The class-conditional velocity network $u_\theta(x,t,y)$ is a residual UNet with two downsampling stages.
The encoder uses $96$, $192$, and $192$ channels at resolutions $32{\times}32$, $16{\times}16$, and $8{\times}8$, respectively; the decoder uses $192$ and $96$ channels at the two higher resolutions, with concatenated skip connections from the encoder.
The input is projected to $96$ channels by a $3{\times}3$ convolution, and the output layer applies group normalization, SiLU, and a $3{\times}3$ convolution to produce the three velocity channels.
Each residual block has two $3{\times}3$ convolutions, each preceded by group normalization with $8$ groups and a SiLU activation; a $1{\times}1$ skip projection is used when the channel count changes.
A linear projection of the conditioning vector is added after the first convolution in each block.
Downsampling uses stride-$2$ $4{\times}4$ convolutions, and upsampling uses transposed convolutions with the same kernel size and stride.
For conditioning, we sum a $192$-dimensional sinusoidal time embedding processed by a two-layer MLP with a SiLU activation and a linear projection of the one-hot class label to the same dimension.

\textbf{Pretraining.}
We minimize the conditional flow-matching loss \eqref{eq:con_fm_loss_2} using $X_t=(1-t)X_0+tX_1$, where $t\sim\mathcal{U}(0,1)$, $X_0\sim\mathcal{N}(0,I)$, and $X_1$ is an independently sampled training image with label $y$.
The regression target is $X_1-X_0$, and the squared error is averaged over image coordinates.
Both pretraining and fine-tuning use AdamW with $(\beta_1,\beta_2)=(0.9,0.999)$, $\epsilon_{\mathrm{Adam}}=10^{-8}$, and gradient-norm clipping at $1.0$.
Pretraining uses batch size $192$, learning rate $2{\times}10^{-4}$, weight decay $10^{-4}$, and cosine learning-rate decay over the training run.
We select three checkpoints, Weak, Medium, and Strong, with validation flow-matching losses $0.225$, $0.196$, and $0.184$, respectively.
Strong is taken after $30$ epochs; Weak and Medium are earlier checkpoints of the same architecture.

\textbf{RL Fine-Tuning and Evaluation.}
We fine-tune each base model independently for $80$ updates.
Each update visits all $10$ classes in a new random order, taking one policy-gradient step per class.
Before each step, we save a frozen copy of the current policy and collect $G=32$ class-conditioned SDE rollouts from independent Gaussian initial states.
We group-normalize their terminal rewards to obtain advantages and take one PPO-clipped update with clip ratio $0.2$ and advantage clipping at $3.0$, using one policy epoch per class.
The learning rate is $4{\times}10^{-6}$ with no weight decay.
A per-step velocity-KL penalty with weight $\beta_{\mathrm{KL}}=0.05$ penalizes deviations from the frozen pretrained policy.
Training rollouts use $10$ Euler--Maruyama steps with the \textsc{flow-grpo} noise schedule $\sigma_t=a\sqrt{(1-t)/t}$, $a=0.7$, from noise to data; ODE evaluation uses $20$ Euler steps.
We repeat fine-tuning with $3$ independent seeds and report means and standard deviations.

We estimate the validation flow-matching loss in \cref{fig:FMRL_val}(d) on a fixed class-balanced subset of $500$ images, with $50$ images per class.
This subset is selected once with a fixed seed and reused across all three base models and fine-tuning updates.
For each loss estimate, we draw fresh independent times and Gaussian noise for each image.

\section{Ablation of FMPO-CPPO}
\label{sec:abl}

\begin{figure}[H]

    \centering

    \begin{subfigure}[c]{0.24\linewidth}
      \centering
      \includegraphics[width=\linewidth]{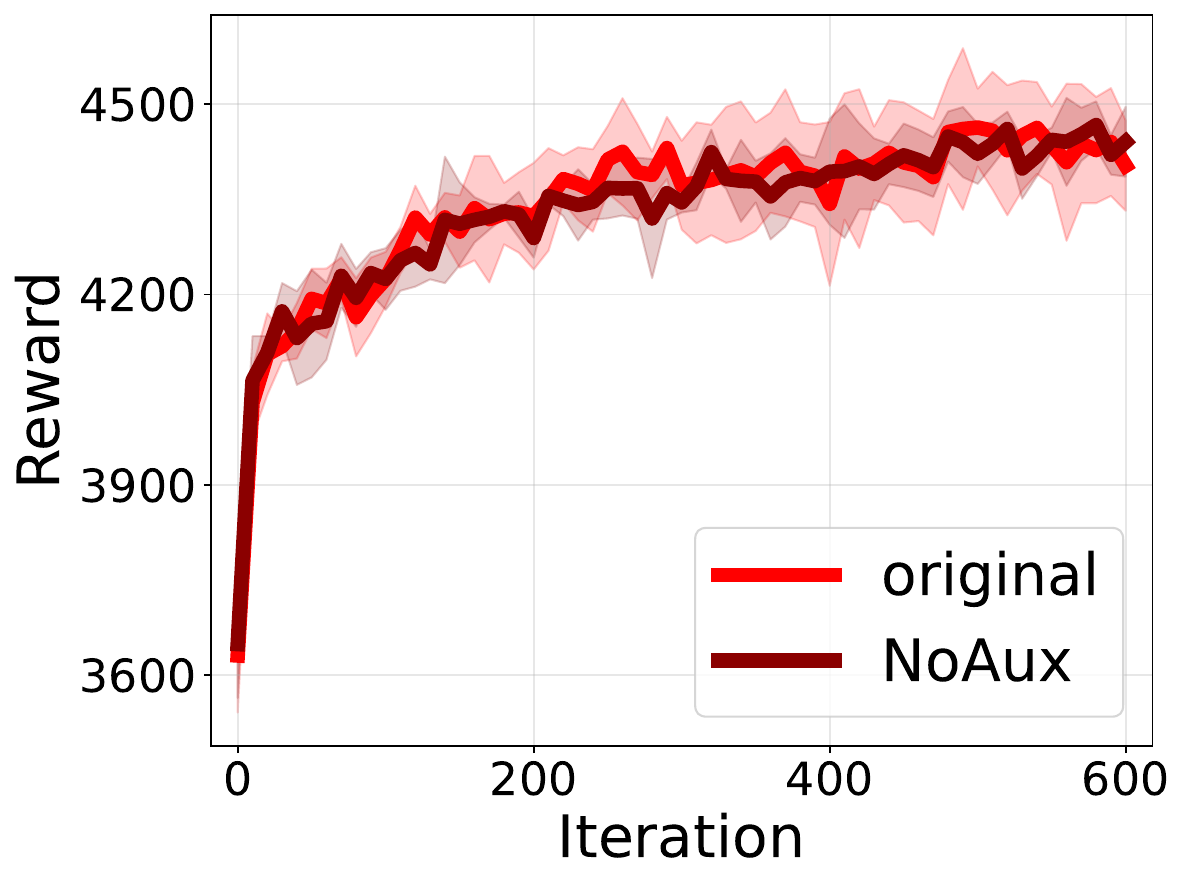}
      \caption*{\small \textbf{(a) Ant-v2}}
    \end{subfigure}\hfill
    \begin{subfigure}[c]{0.24\linewidth}
      \centering
      \includegraphics[width=\linewidth]{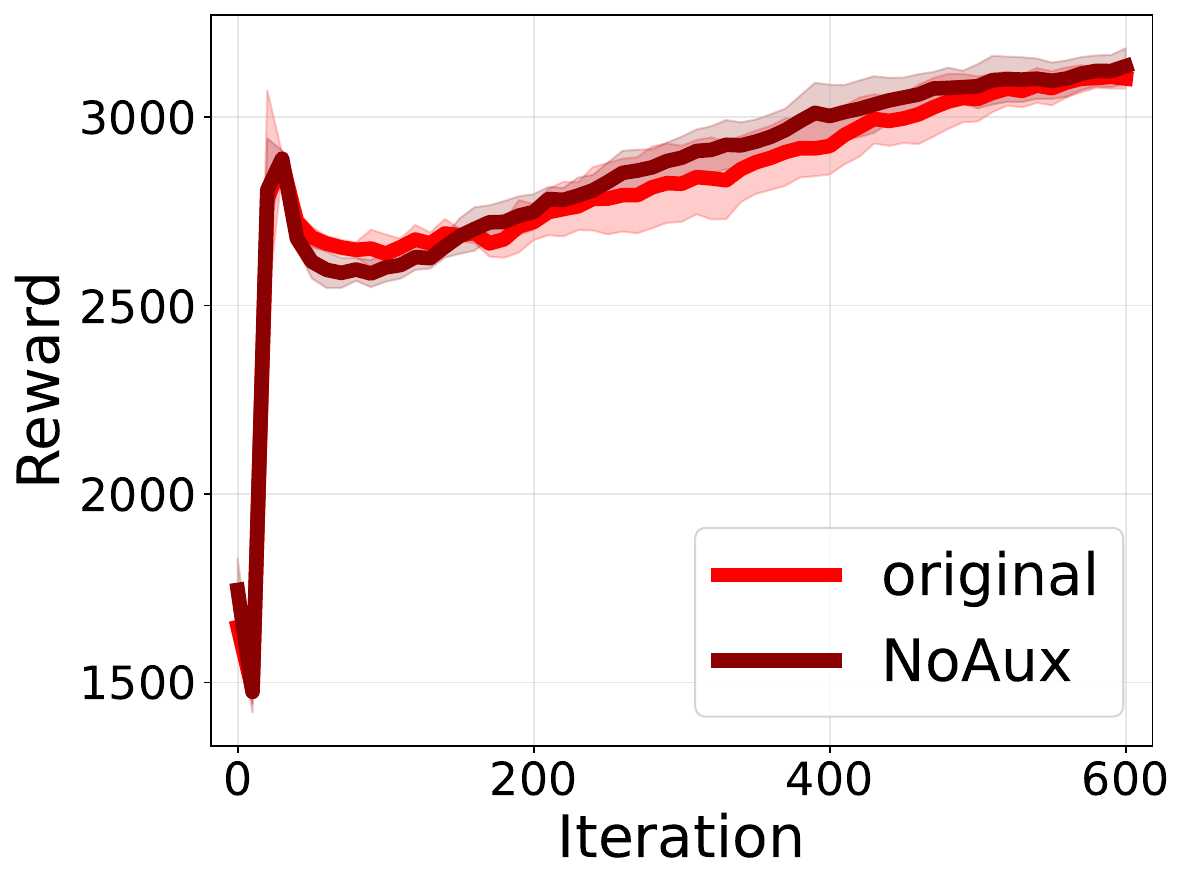}
      \caption*{\small \textbf{(b) Hopper-v2}}
    \end{subfigure}
    \begin{subfigure}[c]{0.24\linewidth}
      \centering
      \includegraphics[width=\linewidth]{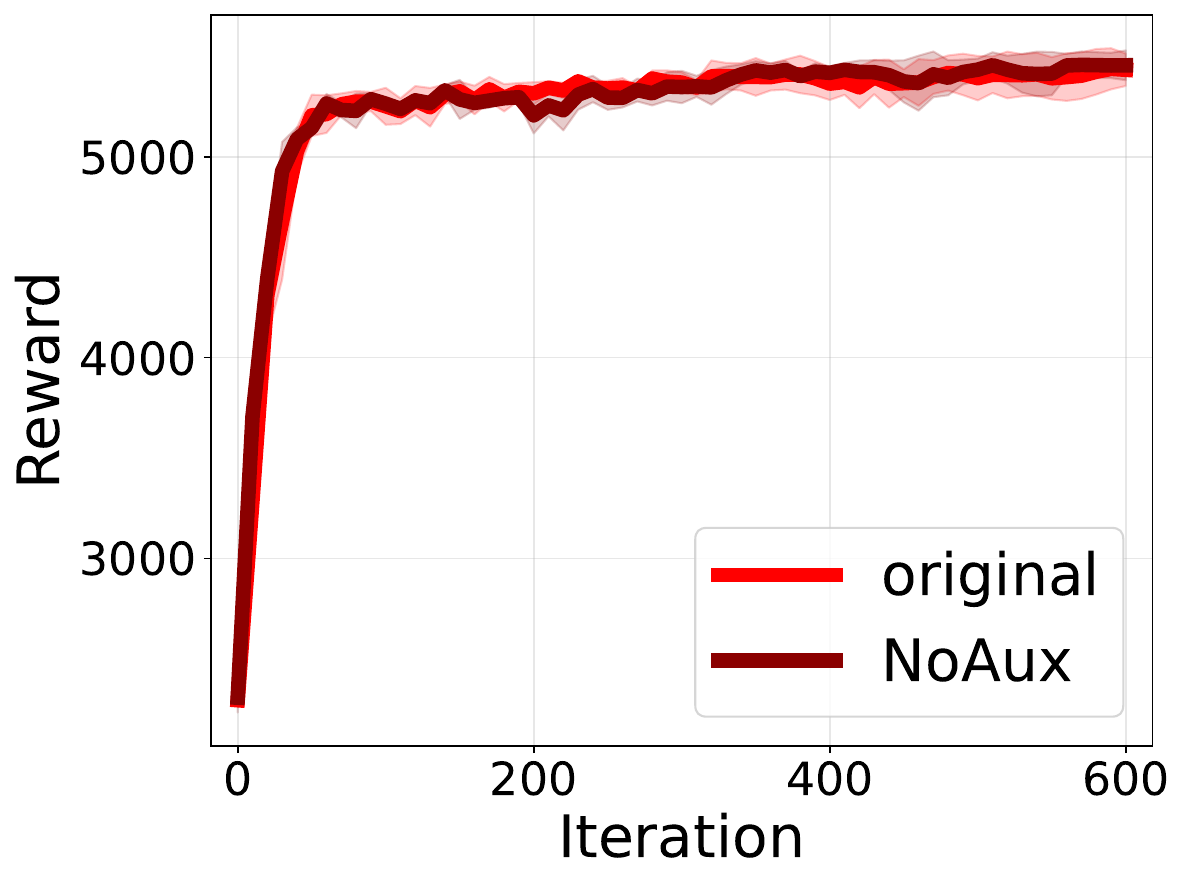}
      \caption*{\small \textbf{(c) Humanoid-v3}}
    \end{subfigure}\hfill
    \begin{subfigure}[c]{0.24\linewidth}
      \centering
      \includegraphics[width=\linewidth]{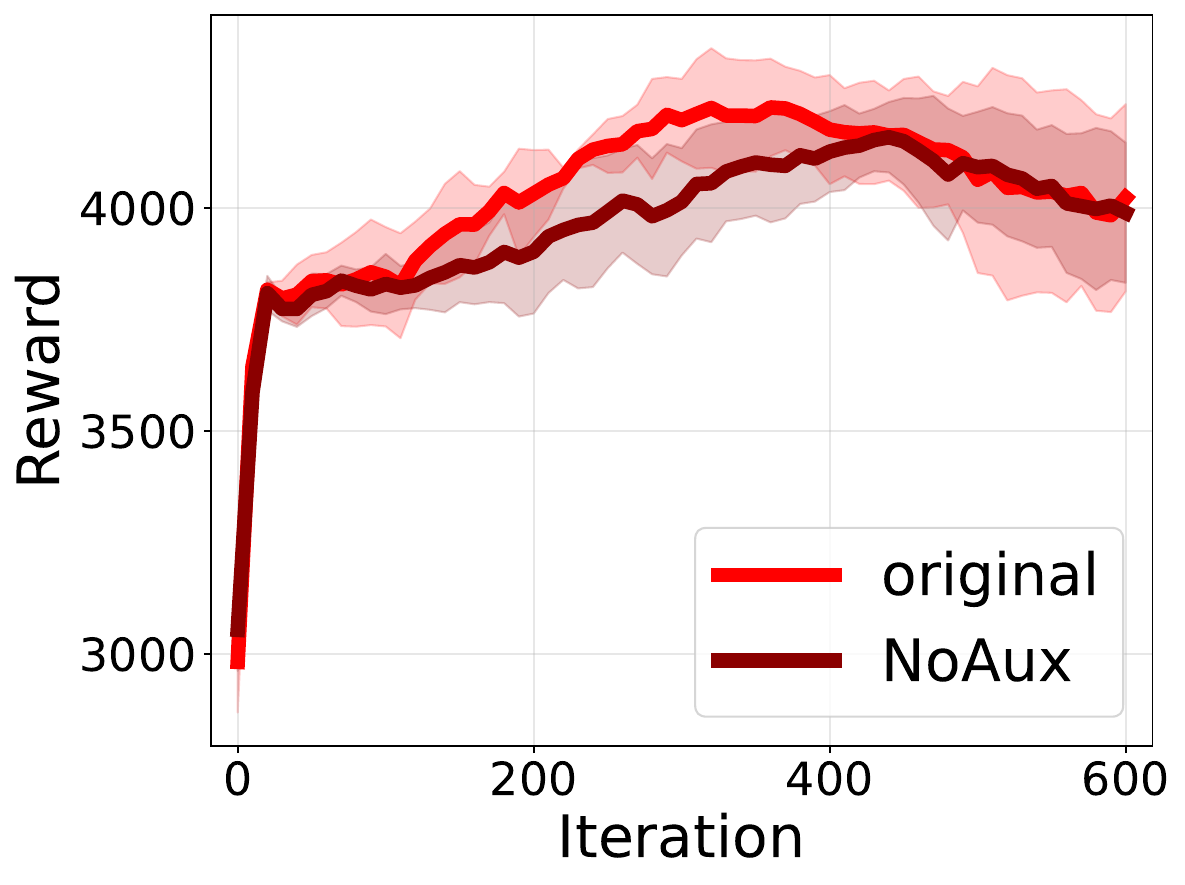}
      \caption*{\small \textbf{(d) Walker-v2}}
    \end{subfigure}
    \caption{Ablation of the auxiliary step-wise PPO term in FMPO-CPPO on four continuous-control tasks.}
    \label{fig:fmpo_auxiliary_ablation}

\end{figure}

We run FMPO-CPPO with the same settings and hyperparameters without the auxiliary term, leaving $\mathcal J_{\rm chain}(\theta)$ as the only optimization objective.
As shown in \cref{fig:fmpo_auxiliary_ablation}, the performance is similar on Ant, Hopper, and Humanoid, while the original FMPO-CPPO performs better on Walker. With only $\mathcal J_{\rm chain}(\theta)$ objective, FMPO-CPPO without the auxiliary term still beats ReinFlow and Flow-SDE with large margins on Ant and Humanoid, while performing similarly on Walker and Hopper.

\stopcontents[appendices]

\clearpage
\phantomsection
\addcontentsline{toc}{section}{References}
\bibliographystyle{plainnat}
\bibliography{refs}

@book{karatzas2014brownian,
  title={Brownian motion and stochastic calculus},
  author={Karatzas, Ioannis and Shreve, Steven},
  year={2014},
  publisher={springer}
}

@article{albergo2022building,
  title={Building normalizing flows with stochastic interpolants},
  author={Albergo, Michael S and Vanden-Eijnden, Eric},
  journal={arXiv preprint arXiv:2209.15571},
  year={2022}
}

@article{le2023voicebox,
  title={Voicebox: Text-guided multilingual universal speech generation at scale},
  author={Le, Matthew and Vyas, Apoorv and Shi, Bowen and Karrer, Brian and Sari, Leda and Moritz, Rashel and Williamson, Mary and Manohar, Vimal and Adi, Yossi and Mahadeokar, Jay and others},
  journal={Advances in neural information processing systems},
  volume={36},
  pages={14005--14034},
  year={2023}
}

@article{black2024pi,
  title={$\pi_0$: A Vision-Language-Action Flow Model for General Robot Control},
  author={Black, Kevin and Brown, Noah and Driess, Danny and Esmail, Adnan and Equi, Michael and Finn, Chelsea and Fusai, Niccolo and Groom, Lachy and Hausman, Karol and Ichter, Brian and others},
  journal={CoRR},
  year={2024}
}

@article{intelligence2025pi5,
  title={$\pi_{0.5}$: a Vision-Language-Action Model with Open-World Generalization},
  author={Intelligence, Physical and Black, Kevin and Brown, Noah and Darpinian, James and Dhabalia, Karan and Driess, Danny and Esmail, Adnan and Equi, Michael and Finn, Chelsea and Fusai, Niccolo and others},
  journal={arXiv preprint arXiv:2504.16054},
  year={2025}
}

@article{park2025flow,
  title={Flow q-learning},
  author={Park, Seohong and Li, Qiyang and Levine, Sergey},
  journal={arXiv preprint arXiv:2502.02538},
  year={2025}
}

@article{mcallister2025flow,
  title={Flow matching policy gradients},
  author={McAllister, David and Ge, Songwei and Yi, Brent and Kim, Chung Min and Weber, Ethan and Choi, Hongsuk and Feng, Haiwen and Kanazawa, Angjoo},
  journal={arXiv preprint arXiv:2507.21053},
  year={2025}
}

@article{li2025mixgrpo,
  title={Mixgrpo: Unlocking flow-based grpo efficiency with mixed ode-sde},
  author={Li, Junzhe and Cui, Yutao and Huang, Tao and Ma, Yinping and Fan, Chun and Yang, Miles and Zhong, Zhao},
  journal={arXiv preprint arXiv:2507.21802},
  year={2025}
}

@article{he2025tempflow,
  title={Tempflow-grpo: When timing matters for grpo in flow models},
  author={He, Xiaoxuan and Fu, Siming and Zhao, Yuke and Li, Wanli and Yang, Jian and Yin, Dacheng and Rao, Fengyun and Zhang, Bo},
  journal={arXiv preprint arXiv:2508.04324},
  year={2025}
}

@article{shao2024deepseekmath,
  title={Deepseekmath: Pushing the limits of mathematical reasoning in open language models},
  author={Shao, Zhihong and Wang, Peiyi and Zhu, Qihao and Xu, Runxin and Song, Junxiao and Bi, Xiao and Zhang, Haowei and Zhang, Mingchuan and Li, YK and Wu, Yang and others},
  journal={arXiv preprint arXiv:2402.03300},
  year={2024}
}

@inproceedings{fan2025online,
  title={Online reward-weighted fine-tuning of flow matching with wasserstein regularization},
  author={Fan, Jiajun and Shen, Shuaike and Cheng, Chaoran and Chen, Yuxin and Liang, Chumeng and Liu, Ge},
  booktitle={The Thirteenth International Conference on Learning Representations},
  year={2025}
}

@article{polyak2025moviegencastmedia,
  title={Movie Gen: A Cast of Media Foundation Models},
  author={Polyak, Adam and Zohar, Amit and Brown, Andrew and Tjandra, Andros and Sinha, Animesh and Lee, Ann and Vyas, Apoorv and Shi, Bowen and Ma, Chih-Yao and Chuang, Ching-Yao and others},
  journal={CoRR},
  year={2024}
}

@inproceedings{esser2024scaling,
  title={Scaling rectified flow transformers for high-resolution image synthesis},
  author={Esser, Patrick and Kulal, Sumith and Blattmann, Andreas and Entezari, Rahim and M{\"u}ller, Jonas and Saini, Harry and Levi, Yam and Lorenz, Dominik and Sauer, Axel and Boesel, Frederic and others},
  booktitle={Forty-first international conference on machine learning},
  year={2024}
}

@article{liu2022flow,
  title={Flow straight and fast: Learning to generate and transfer data with rectified flow},
  author={Liu, Xingchao and Gong, Chengyue and Liu, Qiang},
  journal={arXiv preprint arXiv:2209.03003},
  year={2022}
}

@article{intelligence2025pi,
  title={$\pi_{0.6}^*$: a VLA That Learns From Experience},
  author={Intelligence, Physical and Amin, Ali and Aniceto, Raichelle and Balakrishna, Ashwin and Black, Kevin and Conley, Ken and Connors, Grace and Darpinian, James and Dhabalia, Karan and DiCarlo, Jared and others},
  journal={arXiv preprint arXiv:2511.14759},
  year={2025}
}

@article{wagenmaker2025steering,
  title={Steering Your Diffusion Policy with Latent Space Reinforcement Learning},
  author={Wagenmaker, Andrew and Nakamoto, Mitsuhiko and Zhang, Yunchu and Park, Seohong and Yagoub, Waleed and Nagabandi, Anusha and Gupta, Abhishek and Levine, Sergey},
  journal={arXiv preprint arXiv:2506.15799},
  year={2025}
}

@article{xue2025dancegrpo,
  title={DanceGRPO: Unleashing GRPO on Visual Generation},
  author={Xue, Zeyue and Wu, Jie and Gao, Yu and Kong, Fangyuan and Zhu, Lingting and Chen, Mengzhao and Liu, Zhiheng and Liu, Wei and Guo, Qiushan and Huang, Weilin and others},
  journal={arXiv preprint arXiv:2505.07818},
  year={2025}
}

@article{schulman2017proximal,
  title={Proximal policy optimization algorithms},
  author={Schulman, John and Wolski, Filip and Dhariwal, Prafulla and Radford, Alec and Klimov, Oleg},
  journal={arXiv preprint arXiv:1707.06347},
  year={2017}
}

@article{williams1992simple,
  title={Simple statistical gradient-following algorithms for connectionist reinforcement learning},
  author={Williams, Ronald J},
  journal={Machine learning},
  volume={8},
  number={3},
  pages={229--256},
  year={1992},
  publisher={Springer}
}

@article{lipman2022flow,
  title={Flow matching for generative modeling},
  author={Lipman, Yaron and Chen, Ricky TQ and Ben-Hamu, Heli and Nickel, Maximilian and Le, Matt},
  journal={arXiv preprint arXiv:2210.02747},
  year={2022}
}

@article{su2025flow,
  title={On Flow Matching KL Divergence},
  author={Su, Maojiang and Hu, Jerry Yao-Chieh and Pi, Sophia and Liu, Han},
  journal={arXiv preprint arXiv:2511.05480},
  year={2025}
}

@article{albergo2023stochastic,
  title={Stochastic interpolants: A unifying framework for flows and diffusions},
  author={Albergo, Michael S and Boffi, Nicholas M and Vanden-Eijnden, Eric},
  journal={arXiv preprint arXiv:2303.08797},
  year={2023}
}

@article{liu2025flow,
  title={Flow-grpo: Training flow matching models via online rl},
  author={Liu, Jie and Liu, Gongye and Liang, Jiajun and Li, Yangguang and Liu, Jiaheng and Wang, Xintao and Wan, Pengfei and Zhang, Di and Ouyang, Wanli},
  journal={arXiv preprint arXiv:2505.05470},
  year={2025}
}

@article{black2023training,
  title={Training diffusion models with reinforcement learning},
  author={Black, Kevin and Janner, Michael and Du, Yilun and Kostrikov, Ilya and Levine, Sergey},
  journal={arXiv preprint arXiv:2305.13301},
  year={2023}
}

@article{fan2023optimizing,
  title={Optimizing ddpm sampling with shortcut fine-tuning},
  author={Fan, Ying and Lee, Kangwook},
  journal={arXiv preprint arXiv:2301.13362},
  year={2023}
}

@article{lipman2024flow,
  title={Flow matching guide and code},
  author={Lipman, Yaron and Havasi, Marton and Holderrieth, Peter and Shaul, Neta and Le, Matt and Karrer, Brian and Chen, Ricky TQ and Lopez-Paz, David and Ben-Hamu, Heli and Gat, Itai},
  journal={arXiv preprint arXiv:2412.06264},
  year={2024}
}

@article{ren2024diffusion,
  title={Diffusion policy policy optimization},
  author={Ren, Allen Z and Lidard, Justin and Ankile, Lars L and Simeonov, Anthony and Agrawal, Pulkit and Majumdar, Anirudha and Burchfiel, Benjamin and Dai, Hongkai and Simchowitz, Max},
  journal={arXiv preprint arXiv:2409.00588},
  year={2024}
}

@article{zhang2025reinflow,
  title={ReinFlow: Fine-tuning flow matching policy with online reinforcement learning},
  author={Zhang, Tonghe and Yu, Chao and Su, Sichang and Wang, Yu},
  journal={arXiv preprint arXiv:2505.22094},
  year={2025}
}

@article{brockman2016openai,
  title = {OpenAI Gym},
  author = {Brockman, Greg and Cheung, Vicki and Pettersson, Ludwig and Schneider, Jonas and Schulman, John and Tang, Jie and Zaremba, Wojciech},
  journal = {arXiv preprint arXiv:1606.01540},
  year = {2016}
}

@article{li2023towards,
  title={Towards faster non-asymptotic convergence for diffusion-based generative models},
  author={Li, Gen and Wei, Yuting and Chen, Yuxin and Chi, Yuejie},
  journal={arXiv preprint arXiv:2306.09251},
  year={2023}
}

@article{benton2023error,
  title={Error bounds for flow matching methods},
  author={Benton, Joe and Deligiannidis, George and Doucet, Arnaud},
  journal={arXiv preprint arXiv:2305.16860},
  year={2023}
}

@article{chen2025pi,
  title={$\pi_{RL}$: Online RL Fine-tuning for Flow-based Vision-Language-Action Models},
  author={Chen, Kang and Liu, Zhihao and Zhang, Tonghe and Guo, Zhen and Xu, Si and Lin, Hao and Zang, Hongzhi and Li, Xiang and Zhang, Quanlu and Yu, Zhaofei and others},
  journal={arXiv preprint arXiv:2510.25889},
  year={2025}
}
\end{document}